\documentclass[letterpaper, 10 pt, conference]{ieeeconf}
\IEEEoverridecommandlockouts 
\usepackage{graphics} % for pdf, bitmapped graphics files
\usepackage{epsfig} % for postscript graphics files
\usepackage{amsmath}
\usepackage{amssymb}  % assumes amsmath package installed
\usepackage{bm}  % assumes amsmath package installed
\usepackage{makecell}
\usepackage[noadjust]{cite} % 'sort' and 'compress' are enabled by default
\usepackage{url}
\usepackage[hidelinks]{hyperref}
\usepackage{booktabs} % For tables
\usepackage{subcaption} % For subfigures
\usepackage{graphicx}
\usepackage{textcomp}
\usepackage{float}
\usepackage{lipsum}
\usepackage{algorithm}
\usepackage{algpseudocode}
\usepackage{setspace}
\usepackage{optidef}
\usepackage{caption}
\usepackage{xcolor}
\usepackage{float}
\usepackage{multirow} %for split tables
\usepackage{geometry}
\newgeometry{top=60pt, bottom=43pt, left=48pt, right=48pt}
\title{Chance-Constrained Convex MPC for Robust Quadruped Locomotion\\Under Parametric and Additive Uncertainties}
\author{Ananya Trivedi$^{1}$, Sarvesh Prajapati$^{1}$, Mark Zolotas$^{1,2}$, Michael Everett$^{1}$ and Ta\c{s}k{\i}n Pad{\i}r$^{1,3}$% <-this % stops a space
\thanks{$^{1}$Institute for Experiential Robotics, Northeastern University, Boston, Massachusetts, USA.}
\thanks{$^{2}$Mark Zolotas is currently at Toyota Research Institute (TRI), Cambridge, MA, USA. This paper describes work performed at Northeastern University and is not associated with TRI.}
\thanks{$^{3}$Ta\c{s}k{\i}n Pad{\i}r holds concurrent appointments as a Professor of Electrical and Computer Engineering at Northeastern University and as an Amazon Scholar. This paper describes work performed at Northeastern University and is not associated with Amazon.}
\thanks{Corresponding author: \tt\small trivedi.ana@northeastern.edu}
}
\begin{document}
\maketitle

\begin{abstract}
Recent advances in quadrupedal locomotion have focused on improving stability and performance across diverse environments. However, existing methods often lack adequate safety analysis and struggle to adapt to varying payloads and complex terrains, typically requiring extensive tuning. To overcome these challenges, we propose a Chance-Constrained Model Predictive Control (CCMPC) framework that explicitly models payload and terrain variability as distributions of parametric and additive disturbances within the single rigid body dynamics (SRBD) model. Our approach ensures safe and consistent performance under uncertain dynamics by expressing the model's friction cone constraints, which define the feasible set of ground reaction forces, as chance constraints. Moreover, we solve the resulting stochastic control problem using a computationally efficient quadratic programming formulation. Extensive Monte Carlo simulations of quadrupedal locomotion across varying payloads and complex terrains demonstrate that CCMPC significantly outperforms two competitive benchmarks: Linear MPC (LMPC) and MPC with hand-tuned safety margins to maintain stability, reduce foot slippage, and track the center of mass. Hardware experiments on the Unitree Go1 robot show successful locomotion across various indoor and outdoor terrains with unknown loads exceeding 50\% of the robot's body weight, despite no additional parameter tuning. A video of the results and accompanying code can be found at: \href{https://cc-mpc.github.io/}{https://cc-mpc.github.io/}.
\end{abstract}

\section{Introduction}
\begin{figure} [t]
     \centering
     \begin{subfigure}{0.95\linewidth}
         \centering
         \includegraphics[width=\textwidth]{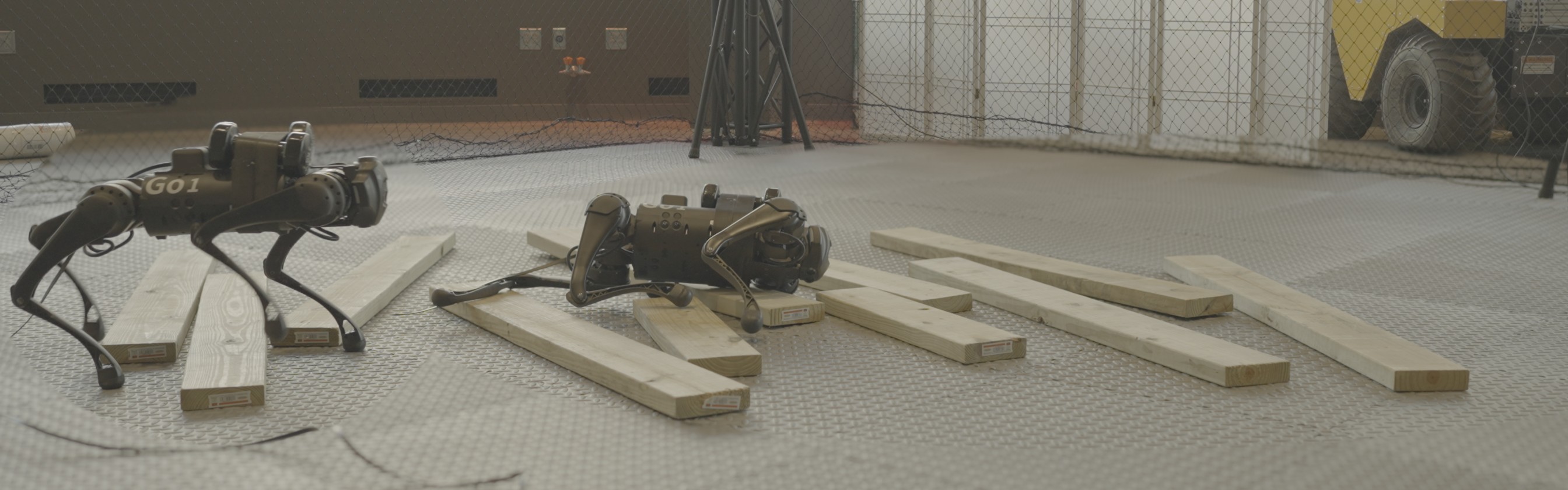}
         \label{fig:y equals x}
     \end{subfigure}
     \begin{subfigure}{0.95\linewidth}
         \centering
         \includegraphics[width=\textwidth]
         {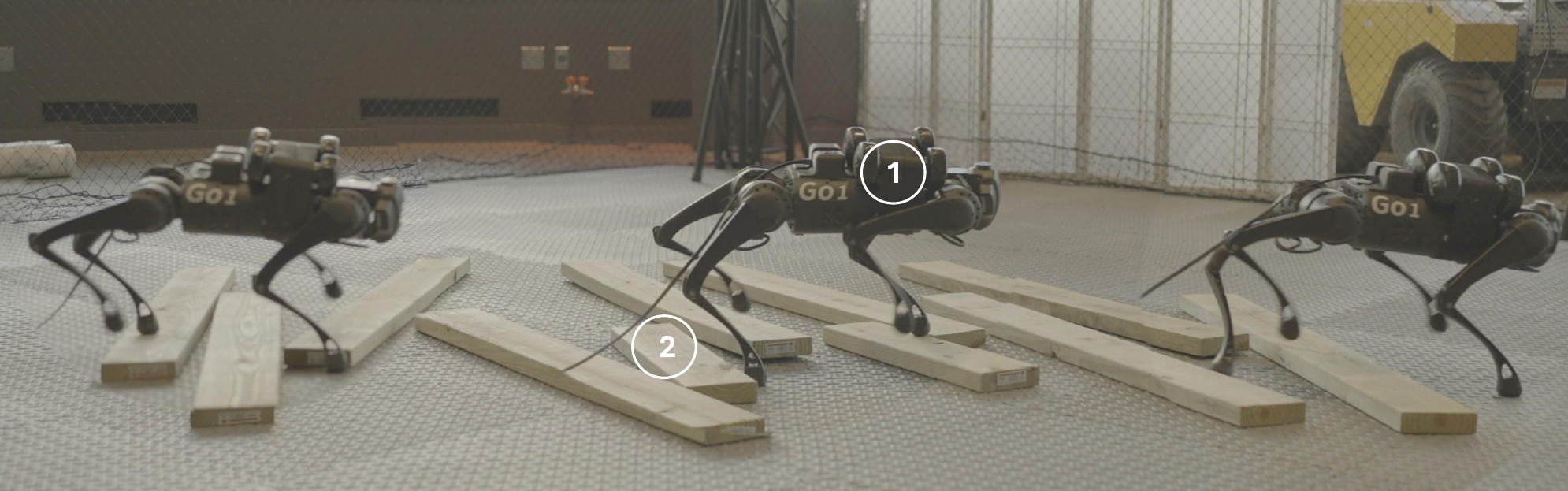}
         \label{fig:three sin x}
     \end{subfigure}
     \caption{Chance-Constrained MPC (bottom) stabilizes the robot by handling a distribution of inertial uncertainties from 6 kg dumbbells (1) and contact uncertainties, from planks (2). Linear MPC (top) fails under these conditions.}
 \label{fig:visual_only_hardware_comparison}
 \vspace{-1.1mm}
\end{figure}

Quadrupedal robots have demonstrated significant potential in various industrial applications and search and rescue missions. These robots enhance productivity by transporting heavy loads and traversing diverse terrains~\cite{quadrupedal_robots_survey}. However, preventing falls in dynamic environments remains a critical challenge~\cite{related_works_rl_marco_hutter}. Inaccurate system models, external disturbances, and unpredictable payload variations can cause deviations from planned motions, resulting in unintended contact locations or timing errors~\cite{related_works_gazar_wholebody}. As shown in Fig.~\ref{fig:visual_only_hardware_comparison}, classical Model Predictive Control (MPC) methods often struggle with these discrepancies, leading to instability, foot slippage, or even falls.

Conventional model-based control methods for quadrupedal locomotion either fail to account for dynamics uncertainties, as in Differential Dynamic Programming (DDP)~\cite{what_is_ddp}, or become computationally infeasible when doing so, as with Stochastic Linear Complementarity Problems (SLCPs)~\cite{cito_1}. Model-free Reinforcement Learning (RL) aims to generalize robot locomotion strategies across diverse environments through offline training, followed by online deployment. While these neural network policies work well in practice, they often lack interpretability and may require frequent re-training to ensure reliable deployment~\cite{rl_parameter_tuning_is_hard}.

In contrast, Stochastic Model Predictive Control (SMPC) directly incorporates uncertainties into the control design by modeling them as probability distributions of disturbances~\cite{stochastic_programming,smpc_survey_paper}. Unlike deterministic MPC, SMPC permits a small probability of constraint violation. In our experiments, we set this probability to 5\%, corresponding to a \(2\sigma\) confidence level. As a result, 95\% of outcomes are expected to remain within constraints under Gaussian disturbances~\cite{gpmpc, related_works_chance_constraints_stanford}. This formulation allows the controller to balance conflicting objectives, such as following a desired trajectory while mitigating unstable behaviors across a range of real-world terrain disturbances.

In this work, we propose a novel Chance-Constrained MPC (CCMPC) algorithm—a specific form of SMPC—to generate ground reaction forces for quadrupedal robots. Our approach models mass, inertia, and contact sequences as stochastic variables. By formulating the control problem as a quadratic programming (QP) problem~\cite{nocedal_and_wright}, we achieve real-time solve rates at $\sim$500 Hz, comparable to Linear MPC (LMPC). We validate CCMPC through extensive simulations and hardware
\newgeometry{top=57pt, bottom=43pt, left=48pt, right=48pt}\hspace*{-9pt}experiments on the Unitree Go1 robot. Our approach achieves superior performance over traditional methods in maintaining stability, reducing foot slippage, and supporting payloads exceeding 50\% of the robot’s weight across muddy slopes, stairs, grass, and gravel. This is accomplished using a unified control policy that effectively handles different terrain conditions and payload variations without the need for parameter tuning. The key contributions of this paper are summarized as follows:
\begin{itemize}
    \item  We develop a CCMPC algorithm tailored for quadrupedal robots to handle disturbances from variable payloads and complex terrain dynamics.
    \item The control problem is formulated as a quadratic program, achieving fast solve times suitable for real-time application.
    \item We validate our method through simulations and hardware experiments. In simulations, CCMPC achieves a 100\% success rate across multiple gaits, compared to 39.2\% for Linear MPC (LMPC) and 75.7\% for hand-tuned MPC. To our knowledge, this is the first SMPC implementation for quadrupedal robots on hardware.

\end{itemize}

\section{Related Works}
Recent advances in trajectory optimization have improved the reliability of quadrupedal locomotion, but managing real-world uncertainties remains challenging. Using convex MPC with the single rigid body dynamics (SRBD) model has enabled fast computation of diverse walking gaits for quadrupedal robots~\cite{mit_nominal_mpc}. These approaches require an accurate dynamics model, making them less effective when the real-world physics deviates from the designed controller model~\cite{related_works_robust_tro}. Prior work has used DDP to account for uncertainties in trajectory optimization for legged locomotion~\cite{related_works_ddp_2}. Employing DDP handles inequality constraints implicitly, making it difficult to address uncertainty impacts on constraint satisfaction~\cite{ddp_yuval}. In contrast, our method uses chance constraints~\cite{how_should_a_robot_assess_risk} to address this issue explicitly.

Methods for optimizing stochastic contact-implicit trajectories in legged robots, such as Expected Residual Minimization (ERM) and SLCP, are computationally demanding and prone to local minima, making them less suitable for real-time applications~\cite{erm,cito2}. Conversely, our approach reformulates the stochastic optimal control problem into a deterministic convex QP problem, which can be solved at 500 Hz~\cite{qpoases}.

In contrast to model-based control, model-free RL techniques eliminate the need for an accurate robot model by using domain randomization during training~\cite{model_free_learning}. This approach exposes the control policies to a wide range of scenarios, enhancing their robustness to diverse environments~\cite{related_works_rl_koushil}. Nevertheless, these controllers face challenges such as the sim-to-real gap, potentially leading to conservative strategies or failure to handle out-of-distribution disturbances~\cite{sim_to_real_gap}. They are also less intuitive to tune and require more engineering effort compared to model-based approaches~\cite{rl_parameter_tuning_is_hard}. Our method, with minimal parameter tuning, demonstrates strong generalization capabilities, as evidenced by extensive Monte Carlo simulations.

Adaptive MPC techniques serve as a middle ground between model-free RL and classical MPC approaches. Existing methods for quadrupedal locomotion estimate residual model uncertainties offline using simulation data~\cite{related_works_pandala_rmpc_plus_rl} or online using L1 adaptive control~\cite{related_works_adaptive_mpc_quan}. In this instance, proper initialization of estimated parameters is crucial to avoid instability before online model convergence~\cite{adaptive_control_textbook}. Additionally, these methods assume constant uncertainty throughout the MPC horizon. By propagating state and control variance along the MPC horizon, our method captures evolving uncertainty, thereby enhancing prediction accuracy.

Gazar et al.~\cite{gazar_biped_rmpc} achieved bipedal walking in simulation despite random forces applied to the robot's CoM. This was accomplished by using tube MPC to handle additive polytopic uncertainties in the dynamics. However, their approach assumed a constant CoM height and zero angular momentum, limiting movement to planar motion without rotation. Xu et al.~\cite{related_works_robust_tro} addressed this limitation by employing robust min-max MPC for quadrupedal walking. Their method considers bounded disturbances and always plans for the worst-case scenario, resulting in overly conservative behavior~\cite{robust_control_survey}. As a result, they had to assign one set of MPC weights for frictional disturbances and another for payload uncertainties. In contrast, SMPC offers greater flexibility than RMPC by allowing small, user-defined probabilities of constraint violation, which mitigates the issue of conservativeness without sacrificing performance~\cite{smpc_survey_paper}. Despite the advantages of SMPC, existing work on quadrupedal robots is mostly confined to simulations and is  computationally intensive for real-time deployment~\cite{smpc_japan,related_works_gazar_wholebody,related_works_gazar_nonlinear,latest_robust_contact}. In this paper, we overcome this computational challenge through an efficient QP implementation. This allows us to deploy the approach on a hardware platform and demonstrate its real-time processing capabilities without the need for hand-tuning, as required in RMPC.

\section{Chance-Constrained Foot Force MPC}

\begin{figure}[t]
    \centering
    \includegraphics[width=\linewidth]{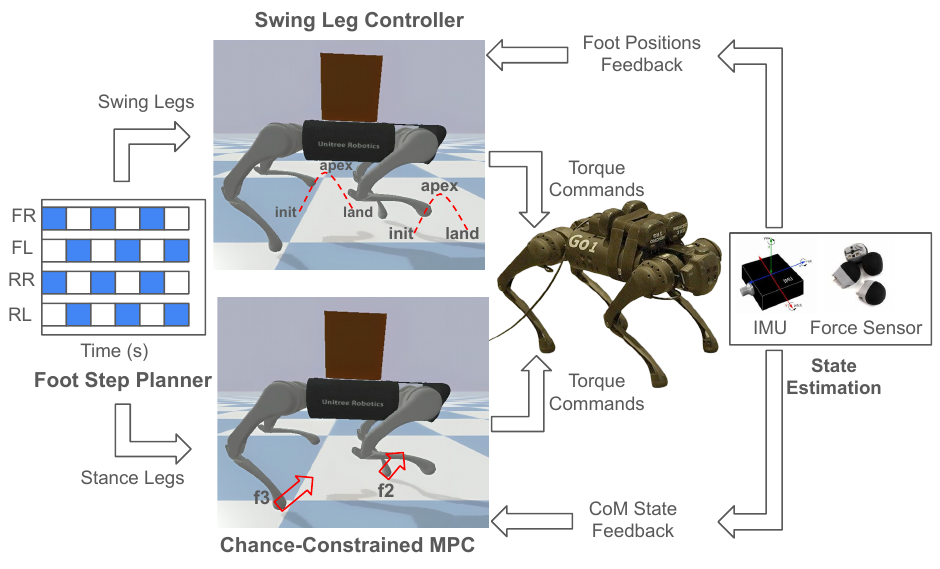}
    \caption{The modular control architecture adopted in this work for quadrupedal locomotion.}
    \label{fig:block_diagram}
    \vspace{-0.4mm}
\end{figure}

In this section, we introduce the Chance-Constrained Foot Force MPC framework, which optimizes ground reaction forces to ensure stability under uncertain conditions. As illustrated in Fig.~\ref{fig:block_diagram}, the framework uses a footstep planner that determines when each leg should enter the swing (white) or stance (blue) phase based on a predefined contact schedule~\cite{fast_and_efficient}. For the swing foot, the Raibert heuristic~\cite{mit_nominal_mpc} calculates the required motor torques, planning the trajectory from initiation through apex to landing using a cubic spline. CCMPC then optimizes ground reaction forces for the stance feet. The state estimator fuses IMU and foot force sensor data using an Extended Kalman Filter (EKF) to estimate the center of mass states and leg end positions.

The whole-body dynamics can be expressed using manipulator equations~\cite{legged_robots_survey}:
\begin{equation} \label{eq:whole_body_dynamics}
\mathbf{M}(\mathbf{q})\ddot{\mathbf{q}} + \mathbf{N}(\mathbf{q}, \dot{\mathbf{q}}) = \mathbf{S}^{T} \boldsymbol{\tau}_{\text{motors}} + \mathbf{J}(\mathbf{q})^{T} \mathbf{f}_{\text{contact}}
\end{equation}
Here, \(\mathbf{q}\) represents the generalized robot pose, \(\mathbf{M}\) is the mass matrix, and \(\mathbf{N}\) encapsulates other nonlinear terms. The selection matrix \(\mathbf{S}\) captures the actuation from the motors, mapping the joint torques to the robot’s dynamics. As discussed in~\cite{nyu}, the foot contact forces \(\mathbf{f}_{\text{contact}}\) can be mapped to desired motor torques \(\boldsymbol{\tau}_{\text{motors}}\) via the contact Jacobian \(\mathbf{J}(\mathbf{q})\). MPC is an effective strategy to determine these foot forces due to its ability to solve constrained optimal control problems~\cite{legged_robots_survey}. Nevertheless, MPC faces challenges when dealing with substantial modeling errors, which can arise from various factors including diverse payloads and terrains. These modeling errors often lead to instability and deviations from desired trajectories~\cite{related_works_robust_tro}.

To address this, we present a stochastic variation of the SRBD model~\cite{mit_nominal_mpc} for quadrupedal robots. The state vector \(\mathbf{x} = [\boldsymbol{\Theta}^T, \mathbf{p}^T, \boldsymbol{\omega}^T, \dot{\mathbf{p}}^T, g]^T\) includes the robot's orientation, position, velocities, and gravity term. The control inputs are the ground reaction forces at each leg, \(\mathbf{u} = [\mathbf{f}_{1}^T, \mathbf{f}_{2}^T, \mathbf{f}_{3}^T, \mathbf{f}_{4}^T]^T \). At time instant \(i\), equations of motion are thus expressed as:
\begin{equation} \label{eq:stochastic_srbd}
\mathbf{x}_{i+1} = \mathbf{A}_i  \mathbf{x}_i + \mathbf{B}_i (\boldsymbol{\delta}_i) \mathbf{u}_i + \mathbf{w}_i,
\end{equation}
\begin{equation*}
\text{where} \quad
\boldsymbol{\delta}_i \sim \mathcal{N}(\mathbb{E}[\boldsymbol{\delta}_i], \boldsymbol{\Sigma}_{\boldsymbol{\delta}}), \quad \mathbf{w}_i \sim \mathcal{N}(\mathbb{E}[\mathbf{w}], \boldsymbol{\Sigma}_{\mathbf{w}})
\end{equation*}
Here, \(\mathbf{A}_i\) is the state transition matrix and \(\mathbf{B}_i(\boldsymbol{\delta}_i)\) is a selection matrix that maps the control inputs to the state, where \(\boldsymbol{\delta}_i\) represents the distribution of parametric uncertainties. Its mean is given by \(\overline{\boldsymbol{\delta}_i} = [m, \text{diag}(\mathbf{I}), \mathbf{r}_{i,1}^T, \ldots, \mathbf{r}_{i,4}^T]\), where \(m\) is the nominal robot mass and \(\text{diag}(\mathbf{I})\) represents the diagonal entries of the nominal inertia matrix. At time instant \(i\), \(\mathbf{r}_{i,1}, \mathbf{r}_{i,2}, \mathbf{r}_{i,3},\) and \(\mathbf{r}_{i,4}\) are the foot locations planned using the Raibert heuristic, as shown in Fig.~\ref{fig:block_diagram}. \(\boldsymbol{\Sigma}_{\boldsymbol{\delta}}\) is the user-tunable covariance matrix representing variations in these parameters due to unknown payloads and uneven terrain. Similarly, \(\mathbb{E}[\mathbf{w}] = \mathbf{0}\) and \(\boldsymbol{\Sigma}_{\mathbf{w}}\) are the mean and covariance matrix of the residual model nonlinearities~\cite{related_works_pandala_rmpc_plus_rl} that are not accounted for by the nominal SRBD model.

We use CCMPC to capture these sources of uncertainty by incorporating probabilistic descriptions of uncertainties into a stochastic optimal control problem~\cite{related_works_chance_constraints_stanford}. Though by modeling uncertainties, guaranteeing constraint satisfaction at all times becomes impractical. Chance constraints instead ensure state and control constraints are satisfied with a specified probability, allowing controlled levels of constraint violation~\cite{related_works_chance_constraints_stanford}. This balance between effective performance and reliability is essential for maintaining consistent operation in uncertain environments.

Specifically, we enforce the linearized friction cone constraints and the unilateral force constraints~\cite{russ_legged_robots} as chance constraints on the ground reaction forces. Mathematically, these constraints can be expressed as:
\begin{equation} \label{eq:chance_constraints}
\Pr(\mathbf{C}_i \mathbf{u}_i \leq \mathbf{0}) \geq \epsilon
\end{equation}
Here, \(\mathbf{C}_i \in \mathbb{R}^{20 \times 12}\) encapsulates the friction cone and unilateral force constraints for the four feet of a quadrupedal robot. Each foot contributes four linearized constraints to approximate the friction cone and one constraint for the unilateral contact force, resulting in a total of 20 constraints. The matrix \(\mathbf{C}_i\) maps these constraints to the 12-dimensional control input vector representing ground reaction forces. The term \(\epsilon\) represents acceptable probability thresholds for constraint satisfaction, allowing a controlled level of constraint violation. Furthermore, we minimize the expected cost, which includes deviations from the desired CoM trajectory and a regularization term on the control inputs to reflect control effort:
\begin{equation} \label{eq:expected_cost}
\mathbb{E}[J(\mathbf{x}, \mathbf{u})] = \mathbb{E}\left[\sum_{i=0}^{N-1} \| \mathbf{x}_{i+1} - \mathbf{x}_{\text{ref},i+1} \|_{\mathbf{Q}}^2 + \| \mathbf{u}_i \|_{\mathbf{R}}^2 \right]
\end{equation}
Here, the expectation is taken with respect to the distributions of the robot state and control actions: \(\mathbf{x} \sim (\bar{\mathbf{x}}, \boldsymbol{\Sigma}_x)\) and \(\mathbf{u} \sim (\mathbf{v}, \boldsymbol{\Sigma}_u)\), where \(\bar{\mathbf{x}}\) and \(\mathbf{v}\) denote the means, and \(\boldsymbol{\Sigma}_x\) and \(\boldsymbol{\Sigma}_u\) denote the covariances. Over the MPC horizon \(N\), the positive semi-definite matrix \(\mathbf{Q}\) weighs the tracking error between the state \(\mathbf{x}\) and the reference state \(\mathbf{x}_{\text{ref}}\), and the positive definite matrix \(\mathbf{R}\) weighs the control effort. Finally, we include a constraint using \(\mathbf{D}_i\) to ensure that forces are zeroed out for legs not in contact with the ground, as determined by the footstep planner:
\begin{equation} \label{eq:no_contact_force}
\mathbf{D}_i \mathbf{u}_i = 0
\end{equation}

The resultant form of the Chance-Constrained Foot Force MPC, incorporating stochastic dynamics and chance inequality constraints, is thus expressed as:
\begin{equation} \label{eq:cc_mpc}
\begin{aligned}
& \underset{\mathbf{x}_i, \mathbf{u}_i}{\text{minimize}} \quad \text{Eqn.~\ref{eq:expected_cost}} \\
& \text{subject to} \quad \text{Eqn.~\ref{eq:stochastic_srbd}}, \; \text{Eqn.~\ref{eq:chance_constraints}},  \; \text{Eqn.~\ref{eq:no_contact_force}}
\end{aligned}
\end{equation}

\section{Convex QP Reformulation of CCMPC}
Solving the original CCMPC problem, as expressed in Eqn.~\ref{eq:cc_mpc}, is computationally infeasible. This is primarily due to the need for integrating multi-dimensional Gaussian probability density functions to evaluate the chance constraints, which becomes intractable for high-dimensional systems~\cite{gpmpc}. In this section, we derive an efficient deterministic reformulation of the stochastic optimal control problem.

\subsection{Uncertainty Propagation}
Due to the parametric uncertainties in the system dynamics and additive disturbances, future predicted states result in a stochastic distribution. Similar to~\cite{related_works_gazar_wholebody}, we parameterize the control law as a state-dependent feedback policy. As a result, future control actions also exhibit uncertainty. Specifically, the control law is of the form:
\begin{equation} \label{eq:control_law}
\mathbf{u}_i = \mathbf{v}_i + \mathbf{K}_i( \mathbf{x}_i - \bar{\mathbf{x}}_i)
\end{equation}
Here, \(\mathbf{v}_i\) is the feedforward control input and \(\bar{\mathbf{x}}_i\) is the predicted mean of the state distribution. The feedforward action, \(\mathbf{v}_i\), provides the primary control effort to achieve the desired trajectory based on the predicted system behavior. The feedback action, \(\mathbf{K}_i( \mathbf{x}_i - \bar{\mathbf{x}}_i)\), adjusts for any deviations from the predicted trajectory due to disturbances or model inaccuracies. Optimizing both the feedback gains \( \mathbf{K}_i \) and the control inputs \(\mathbf{v}_i\) leads to a bi-level optimization problem, which is non-convex and computationally expensive for real-time MPC applications~\cite{bilevel_optimization}. Instead, we precompute the gains \( \mathbf{K}_i \) using efficient solvers for the Discrete Algebraic Riccati Equation (DARE)~\cite{drake}. Thus, the decision variables of our optimization problem are \(\mathbf{v}_i\) and \(\bar{\mathbf{x}}_i\).

Next, we present expressions for the first and second moments of the trajectory distributions. For detailed derivations, interested readers are referred to~\cite{related_works_chance_constraints_stanford}. Applying the expectation operator to Eqn.~\ref{eq:stochastic_srbd}, we obtain:
\begin{equation} \label{eq:mean_dynamics}
\mathbb{E}[\mathbf{x}_{i+1}] = \bar{\mathbf{x}}_{i+1} = \mathbf{A}_i \bar{\mathbf{x}}_i + \mathbf{B}_i (\bar{\boldsymbol{\delta}}_i) \bar{\mathbf{u}}_i
\end{equation}

The covariance of the state distribution at the next time step can be expressed as:
\begin{equation} \label{eq:covariance_dynamics}
\begin{aligned}
\boldsymbol{\Sigma}_{\mathbf{x}_{i+1}} &= \mathbb{E}[(\mathbf{x}_{i+1} - \bar{\mathbf{x}}_{i+1})(\mathbf{x}_{i+1} - \bar{\mathbf{x}}_{i+1})^T] \\
&= \mathbf{A}_{cl} \boldsymbol{\Sigma}_{\mathbf{x}_i} \mathbf{A}_{cl}^T + \mathbf{P}_i \boldsymbol{\Sigma}_{\boldsymbol{\delta}} \mathbf{P}_i^T + \boldsymbol{\Sigma}_{\mathbf{w}}
\end{aligned}
\end{equation}
Here, \(\mathbf{A}_{cl} = \mathbf{A}_i + \mathbf{B}_i (\bar{\boldsymbol{\delta}}_i) \mathbf{K}_i\) represents the closed-loop system dynamics, and \(\mathbf{P}_i\) is the Jacobian of \(\bar{\mathbf{x}}_{i+1}\) with respect to the mean of the parameters \(\bar{\boldsymbol{\delta}}_i\):
\begin{equation} \label{eq:jacobian_P}
\mathbf{P}_i = \frac{\partial \bar{\mathbf{x}}_{i+1}}{\partial \bar{\boldsymbol{\delta}}_i} = \frac{\partial (\mathbf{A}_i \bar{\mathbf{x}}_i + \mathbf{B}_i (\bar{\boldsymbol{\delta}}_i) \bar{\mathbf{u}}_i)}{\partial \bar{\boldsymbol{\delta}}_i}
\end{equation}

Based on these equations, the mean and covariance of the control distribution can be expressed as:
\begin{equation} \label{eq:control_distribution}
\mathbb{E}[\mathbf{u}_{i}] = \mathbf{v}_i \quad \text{and} \quad \boldsymbol{\Sigma}_{\mathbf{u}_{i}} = \mathbf{K}_i \boldsymbol{\Sigma}_{\mathbf{x}_i} \mathbf{K}_i^T
\end{equation}

We now show how using these Gaussian-distributed trajectories allows us to analytically derive a deterministic counterpart of the chance control constraints.

\subsection{Friction Cone Constraint Adjustment}
To ensure there is no slippage between the robot's feet and the ground, the reaction forces must stay within the linearized friction pyramid and satisfy the unilaterality constraint~\cite{legged_robots_survey}. This results in five chance constraints per foot that must be jointly satisfied~\cite{related_works_gazar_wholebody}. The Boole-Bonferroni inequality~\cite{related_works_gazar_nonlinear} allows us to conservatively approximate the probability of satisfying these joint chance constraints by summing the probabilities of each individual constraint for all $n=4$ feet.
\begin{equation} \label{eq:boole}
\sum_{j=1}^{5n} \Pr(\mathbf{C}_i^j \mathbf{u}_i > 0) \leq 1 - \epsilon, \hspace{0.2em} \implies \text{Eqn.~\ref{eq:chance_constraints}}
\end{equation}
where, $\mathbf{C}_i^j$ is the $j^{\text{th}}$ row of the $\mathbf{C}_i$ matrix.

While optimizing risk allocation for each constraint could be more effective, it involves a two-stage optimization problem that can be computationally expensive~\cite{chance_constrained_optimal_risk_allocation}. To circumvent this problem, we assign uniform risks \(\alpha\) to each constraint, where \(\alpha = (1-\epsilon)/5n\). Hence, Eqn.~\ref{eq:boole} becomes:
\begin{equation} \label{eq:uniform_risk_allocation}
\Pr(\mathbf{C}_i^j \mathbf{u}_i > 0) \leq \alpha \implies \Pr(\mathbf{C}_i^j \mathbf{u}_i \leq 0) \leq 1 - \alpha
\end{equation}

Finally, since the control actions follow a normal distribution, \(\mathbf{u}_i \sim \mathcal{N}(\mathbf{v}_i, \mathbf{\Sigma}_i^u)\), individual chance constraints in Eqn.~\ref{eq:uniform_risk_allocation} can be deterministically reformulated as~\cite{gaussin_allows_deterministic}:

\begin{equation} \label{eq:deterministic_reformulation}
\mathbf{C}_i^j \mathbf{v}_i + \phi^{-1}(1-\alpha) \sqrt{\mathbf{C}_i^j \, \mathbf{\Sigma}_i^u \, (\mathbf{C}_i^j)^T} \leq 0
\end{equation}
where \(\phi^{-1}\) is the inverse of the cumulative distribution function of a standard normal distribution~\cite{related_works_gazar_wholebody}. Applying Eqn.~\ref{eq:deterministic_reformulation} to all rows of \(\mathbf{C}_i\), we get the following constraints on the mean of control actions:
\begin{equation} \label{eq:mean_control_constraints}
\mathbf{C}_i \mathbf{v}_i \leq \mathbf{c}_i, \\
\end{equation}
\begin{equation*}
\mathbf{c}_i = \text{col}(c_i^1, \ldots, c_i^{5n}) \, \text{and} \, c_i^k = -\phi^{-1}(1-\alpha)\sqrt{\mathbf{C}_i^k \, \mathbf{\Sigma}_i^u \, (\mathbf{C}_i^k)^T}
\end{equation*}

\begin{figure}[t]
    \centering
    \includegraphics[width=0.83\linewidth]{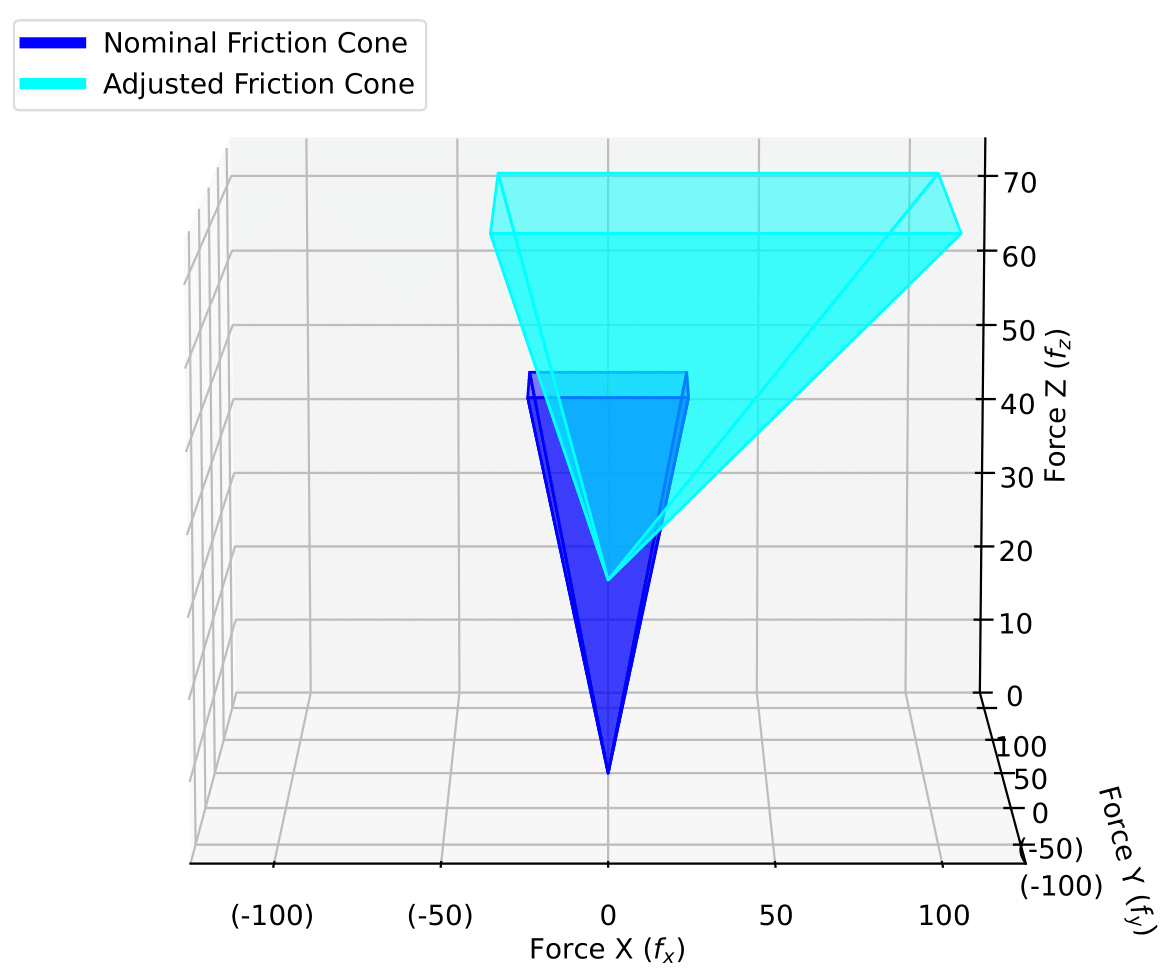}
    \caption{Illustration of friction cone constraint adjustment.}
    \label{fig:friction_cone_vis}
    \vspace{-2.5mm}
\end{figure}

This modification of the original inequality constraints, based on the computed \(\mathbf{c}_i\) values in Eqn.~\ref{eq:mean_control_constraints}, adjusts the feasible set of ground reaction forces to capture variations in operating conditions and robot dynamics~\cite{smpc_survey_paper}. The extent of this adjustment is determined by the constraint satisfaction threshold \(\epsilon\) and the control covariance \(\boldsymbol{\Sigma}_i^{\mathbf{u}}\) propagated along the MPC horizon. Fig.~\ref{fig:friction_cone_vis} illustrates this schematically.

Alternatively, we could have also heuristically chosen a constant value for $\mathbf{c}_i$~\cite{related_works_gazar_wholebody}. In Section~\ref{sec:experiments}, we demonstrate how this heuristic-based approach results in increased constraint violations, leading to greater instability. Despite the additional computation required to evaluate the \(\mathbf{c}_i\) factors for constraint modification, the resultant stance-leg control problem can still be solved efficiently at 500 Hz, making it suitable for real-time implementation.

\subsection{Reduction to Quadratic Program }
In this section, we consolidate the setup of the MPC optimization problem. Following the approach in~\cite{expected_cost}, we use a mean-equivalent approximation of the expected cost along the nominal trajectory:
\begin{equation} \label{eq:mean_equivalent_cost}
\text{Eqn.~\ref{eq:expected_cost}} \approx J(\bar{\mathbf{x}}, \mathbf{v}) = \sum_{i=0}^{N-1} \| \bar{\mathbf{x}}_{i+1} - \mathbf{x}_{\text{ref},i+1} \|_{\mathbf{Q}}^2 + \| \mathbf{v}_i \|_{\mathbf{R}}^2
\end{equation}
Finally, we apply the constraint Eqn.~\ref{eq:no_contact_force} to the mean of the contact forces, that is 
\begin{equation} \label{eq:mean_no_contact_force}
\mathbb{E}[\mathbf{D}_i \mathbf{u}_i] = \mathbf{D}_i \mathbf{v}_i = \boldsymbol{0}
\end{equation}
Friction cone constraint tightening is based on the previous MPC solution, allowing these factors to be pre-computed and fixed for real-time optimization~\cite{gpmpc}.

In summary, given a nominal contact plan and a desired CoM trajectory, CCMPC involves solving the deterministic reformulation of the original stochastic optimization problem as the following quadratic program:
\begin{equation} \label{eq:final_qp}
\begin{aligned}
&\{\bar{\mathbf{x}}^*, \mathbf{v}^*\} = \underset{\bar{\mathbf{x}}_i, \mathbf{v}_i}{\text{minimize}} \quad \text{Eqn.~\ref{eq:mean_equivalent_cost}} \\
&\text{subject to} \quad \text{Eqn.~\ref{eq:mean_dynamics}}, \; \text{Eqn.~\ref{eq:mean_control_constraints}}, \;  \text{Eqn.~\ref{eq:mean_no_contact_force}}
\end{aligned}
\end{equation}
Using single shooting~\cite{mpc_textbook}, we further reduce the number of decision variables in Eqn.~\ref{eq:final_qp} to only include the mean of the control actions. The resultant optimization problem is solved using the qpOASES solver~\cite{qpoases}. The first element of the optimal ground reaction force sequence $\mathbf{v}^*$ is then converted to desired foot motor torques using the contact Jacobian, $\mathbf{J}(\mathbf{q})$. Algorithm~\ref{alg:cc_mpc} outlines the steps necessary to implement the proposed stance-leg controller for quadrupedal locomotion.

\begin{algorithm}[t] \label{algorithm:cc_algorithm}
\caption{Chance-Constrained MPC}
\label{alg:cc_mpc}
\begin{algorithmic}

\State Initialize $\mathbf{c}_i \leftarrow \mathbf{0}$
\While{goal configuration is not reached,}
    \For{$i = 0$ to $N-1$} \Comment{\textcolor{gray}{MPC Loop}}
        \State Compute $\bar{\boldsymbol{\delta}}_i$, $\mathbf{A}_i$, $\mathbf{B}_i (\bar{\boldsymbol{\delta}}_i)$, $\mathbf{D}_i$ using Eqn.~\ref{eq:stochastic_srbd}
        % \State Extract $\mathbf{c}_i$
        \State Add cost function and constraints to Eqn.~\ref{eq:final_qp}
    \EndFor 
    \State \textbf{\textcolor{black}  {Output: }}$\bar{\mathbf{x}}^*, \mathbf{v}^* \leftarrow $ 
    Solve Eqn.~\ref{eq:final_qp}\Comment{\textcolor{gray}{MPC Output}}
    
    \State Initialize $\mathbf{\Sigma_x} \leftarrow \mathbf{0}$\Comment{\textcolor{gray}{Constraint Tightening Loop}}
    \For{$i = 0$ to $N-1$} 
        \State $\mathbf{K}_i \leftarrow \text{DARE}(\mathbf{A}_i,\mathbf{B}_i (\bar{\boldsymbol{\delta}}_i),\mathbf{Q},\mathbf{R})$ 
        \State $\boldsymbol{\Sigma}_{\mathbf{u}} \leftarrow \text{Propagate Control Variance, Eqn.}~\ref{eq:control_distribution}$ 
        \State $\mathbf{c}_i \leftarrow \text{Compute Factors, Eqn.}~\ref{eq:mean_control_constraints}$
        \State $\boldsymbol{\Sigma}_{\mathbf{x}} \leftarrow \text{Propagate State Variance, Eqns.}~\ref{eq:covariance_dynamics},~\ref{eq:jacobian_P}$ 
    \EndFor
    
\EndWhile
\end{algorithmic}
\end{algorithm}

\section {Experiments} \label{sec:experiments}
In this section, we present simulation and hardware experiments evaluating our CCMPC framework. CCMPC effectively tracks the desired height with unmodeled 6 kg payloads and successfully handles loads up to 7.3 kg, exceeding the manufacturer’s recommended payload limit. Extensive indoor and outdoor tests across diverse terrains—including grass, gravel, muddy slopes, stairs, and slippery surfaces—confirm its stability under various payloads. Additional results are available in the supplementary video.

\begin{figure*}[t]
     \centering
     \begin{subfigure}[t]{0.48\textwidth}
         \centering
         \includegraphics[width=0.23\textwidth, height=0.202\textwidth]{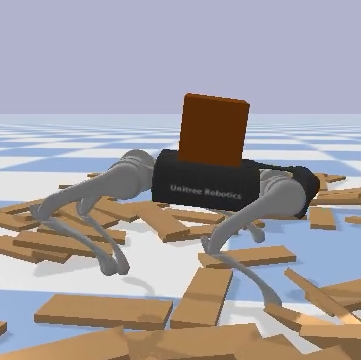}
         \includegraphics[width=0.23\textwidth, height=0.202\textwidth]{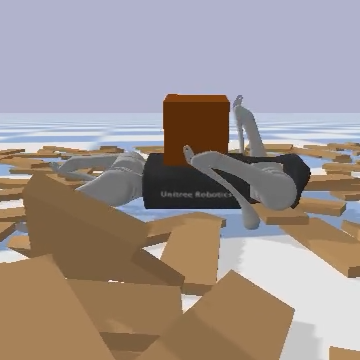}
         \hspace{0.005\textwidth}
         \includegraphics[width=0.23\textwidth, height=0.202\textwidth]{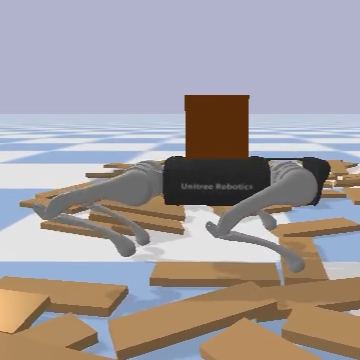}
         \includegraphics[width=0.23\textwidth, height=0.202\textwidth]{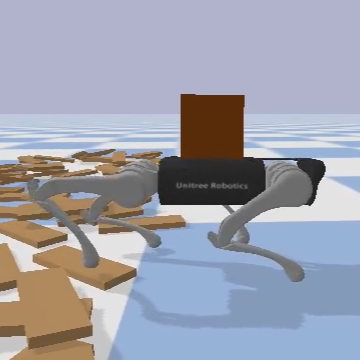}
         \caption{Planks}
         \label{fig:plank_sim}
     \end{subfigure}
     % \hfill
     \begin{subfigure}[t]{0.48\textwidth}
         \centering
         \includegraphics[width=0.23\textwidth]{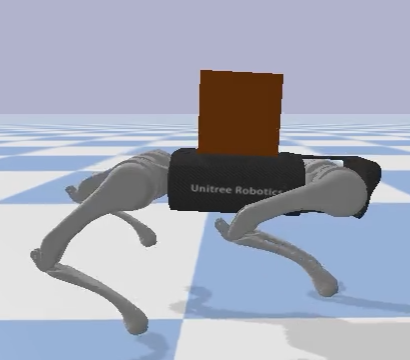}
         \includegraphics[width=0.23\textwidth]{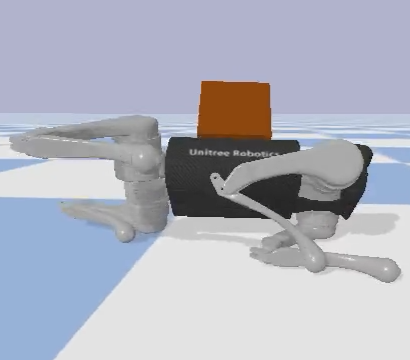}
         \hspace{0.005\textwidth}
         \includegraphics[width=0.23\textwidth]{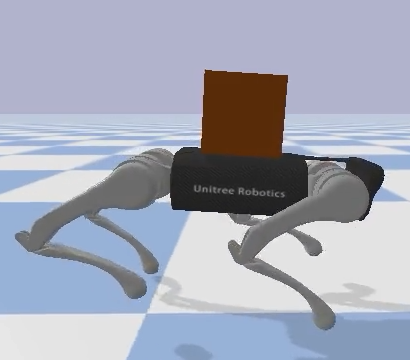}
         \includegraphics[width=0.23\textwidth]{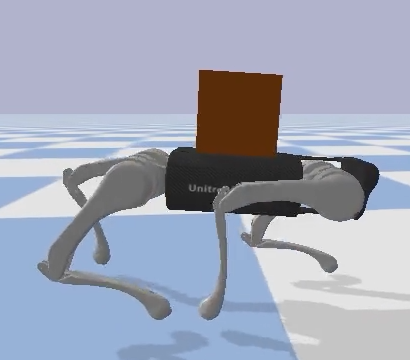}
         \caption{Flytrot Gait}
         \label{fig:fly_trot_sim}
     \end{subfigure}
     \begin{subfigure}[t]{0.48\textwidth}
         \centering
         \includegraphics[width=0.23\textwidth]{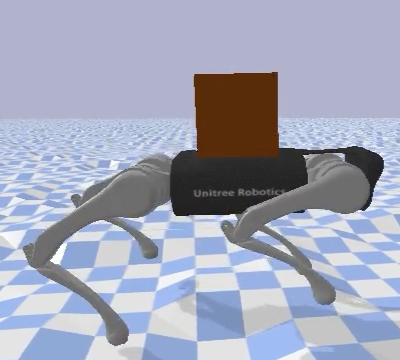}
         \includegraphics[width=0.23\textwidth]{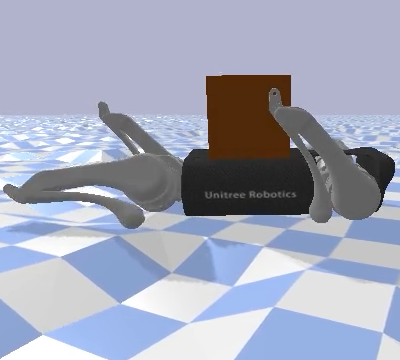}
         \hspace{0.005\textwidth}
         \includegraphics[width=0.23\textwidth]{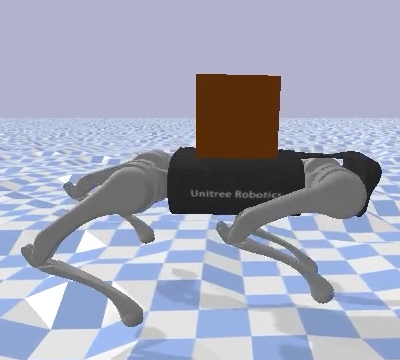}
         \includegraphics[width=0.23\textwidth]{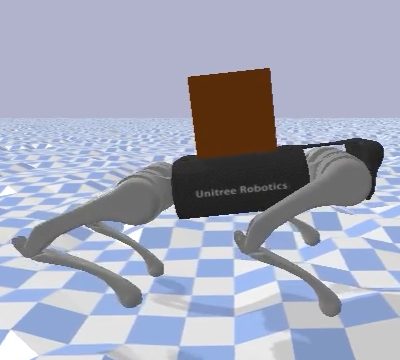}
         \caption{Random Elevation Wavefield}
         \label{fig:uneven_sim}
     \end{subfigure}
     \begin{subfigure}[t]{0.48\textwidth}
         \centering
         \includegraphics[width=0.23\textwidth]{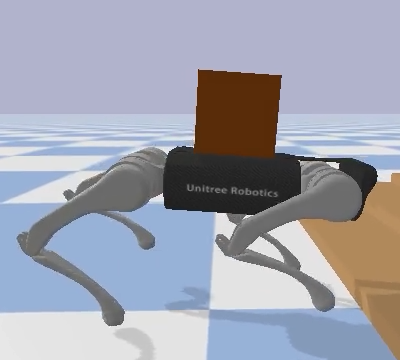}
         \includegraphics[width=0.23\textwidth]{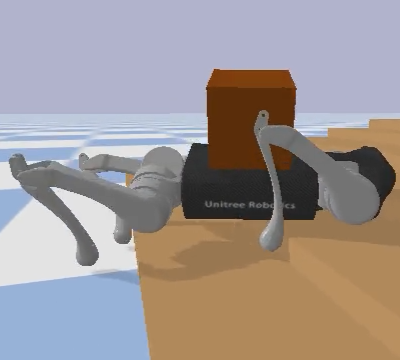}
         \hspace{0.005\textwidth}
         \includegraphics[width=0.23\textwidth]{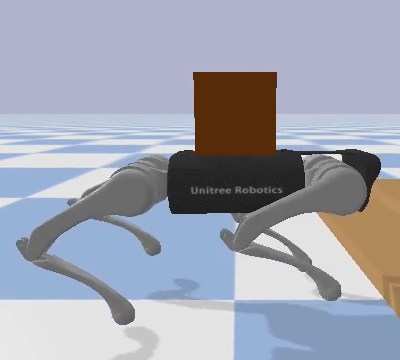}
         \includegraphics[width=0.23\textwidth]{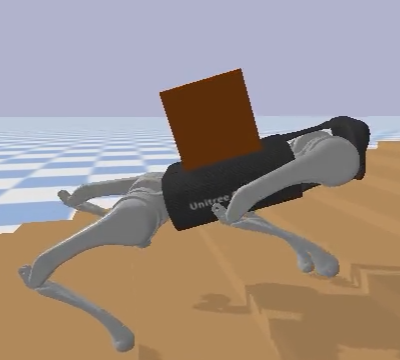}
         \caption{Blind Stair Climbing}
         \label{fig:stair_sim}
     \end{subfigure}

    \caption{Comparison of LMPC (left, fails) and CCMPC (right, succeeds) across various gaits and terrains. Similar failures were also observed with HMPC in these experiments.}
    \label{fig:sim_hard}
    \vspace{-2.3mm}
\end{figure*}

\subsection{Implementation Details}
\begin{table}[t]
\centering
\caption{MPC specific parameters used across all methods.}
\label{tab:mpc_parameters}
\footnotesize
\begin{tabular}{lc}
\toprule
\textbf{Parameter} & \textbf{Value} \\
\midrule
Stepping Frequency & 2.5 Hz \\
Foot Height & 0.08 m \\
Planning Horizon & 10 steps \\
Planning Timestep & 0.025 s \\
Position Weight (Z) & 500 \\
Velocity Weights (X, Y) & 20, 5 \\
Angular Velocity Weights (X, Y, Z) & 0.2, 0.2, 1.0 \\
Roll and Pitch Weights & 0.2, 0.2 \\
Control Penalty Weights & 1e-6 \\
\bottomrule
\end{tabular}
\end{table}

\begin{table}[t]
\centering
\caption{CCMPC variance parameters.}
\label{tab:ccmpc_parameters}
\footnotesize
\begin{tabular}{lc}
\toprule
\textbf{Parameter} & \textbf{Value} \\
\midrule
Mass & 15.0 \\
Inertia (X, Y, Z) & 0.02, 0.06, 0.06 \\
Angular Velocity (X, Y, Z) & 0.5, 0.2, 0.01 \\
Linear Velocity (X, Y, Z) & 0.5, 0.2, 0.01 \\
Contact Location (X, Y, Z) & 0.36, 0.36, 0.36  \\
\bottomrule
\end{tabular}
\vspace{-.2mm}
\end{table}

We validate our approach on the Unitree Go1 robot, which has a total mass of 12 kg. The legs contribute around 10\% of the total weight, allowing us to approximate the robot's dynamics with a SRBD model, as in Eqn.~\ref{eq:stochastic_srbd}~\cite{mit_nominal_mpc}. The robot’s pose and velocities are obtained using onboard IMU and motor encoders. All algorithms are executed on a Legion 5 Pro laptop with an Intel i7-12700H processor and 32 GB RAM. Communication between the robot and laptop is facilitated using Lightweight Communications and Marshalling (LCM)~\cite{lcm}. The reported runtime of 500 Hz accounts for all the computational steps in the control pipeline. This includes the footstep planner, the inverse kinematics-based swing leg controller, the calculation of constraint adjustment factors, and solving the resulting QP.

We compare our CCMPC approach against two benchmarks: Linear MPC (LMPC), which lacks friction cone constraint tightening, and Heuristic MPC (HMPC), which uses hand-computed constraint tightening factors. The MPC-specific parameters used for these three methods are detailed in Table~\ref{tab:mpc_parameters}. The parametric and additive uncertainties for CCMPC, corresponding to 
\(\boldsymbol{\Sigma}_{\boldsymbol{\delta}}\) and  \(\boldsymbol{\Sigma}_{\mathbf{w}}\) in Eqn.~\ref{eq:stochastic_srbd}, are shown in Table~\ref{tab:ccmpc_parameters}. %The constraint satisfaction threshold $\epsilon$ was set to 0.95, ensuring that the chance constraints were respected more than 95\% of the time. -- COMMENTING THIS OUT SINCE IT IS MENTIONED IN INTRODUCTION NOW

\subsection{Simulation Analysis}
The simulation experiments are conducted using the PyBullet simulator~\cite{pybullet} with a high-fidelity dynamics model. We conduct Monte Carlo simulations to demonstrate that our control policy can stabilize quadrupedal motion across various payloads and terrains without parameter adjustments. The robot was commanded to move forward at 0.25 m/s, performing blind locomotion over wooden planks with varying payloads, as shown in Fig.~\ref{fig:sim_hard}(a). The payloads were uniformly sampled within a range of 1 to 10 kg, while the plank heights were uniformly sampled from 0 to 5 cm. A total of 1000 samples were generated to introduce sufficient variability in inertial properties and contact locations. Given the robot’s mass of 12 kg and a foot raise height of 8 cm, these values represent significant variations from nominal conditions. For heuristic constraint tightening, we considered a maximum unmodeled payload of 10 kg and calculated the necessary gravity compensation in the Z direction from the stance feet. To accommodate the maximum commanded acceleration of 0.2 m/s$^2$, we ensured the robot could generate sufficient force in both the X and Y directions, even with the additional payload. This method allowed us to heuristically determine the necessary constraint tightening factors for all directions. 

\begin{table}[t]
    \caption{Results from 1000 Monte-Carlo Simulations.}
    \label{tab:performance_comparison}
    \centering
    \footnotesize
    \setlength\tabcolsep{3pt} % Reduced spacing between columns
    \begin{tabular}{cccccc}
        \toprule
        \textbf{Metric} & \textbf{Gait} & \textbf{LMPC} & \textbf{HMPC} & \textbf{CCMPC} \\
        \midrule
        \multirow{2}{*}{\textbf{Success Rate (\%)} $\uparrow$} & Trot & 38.1 & 77.7 & \textbf{100} \\
         & Flytrot & 40.3 & 73.7 & \textbf{100} \\
        \midrule
        \multirow{2}{*}{\textbf{Average Slippage Ratio} $\downarrow$} & Trot & 0.26 & 0.12  & \textbf{0.11}  \\
         & Flytrot & 0.25  & 0.17  & \textbf{0.15}  \\
        \midrule
        \multirow{2}{*}{\textbf{Normalized Tracking Cost} $\downarrow$} & Trot & $1.4\!\pm\!0.1$ & $1.1\!\pm\!0.2$ & \textbf{1.0$\!\pm\!$0.1} \\
         & Flytrot & $2.2\!\pm\!0.4$ & $1.9\!\pm\!0.2$ & \textbf{1.0$\!\pm\!$0.3} \\
        \midrule
        \multirow{2}{*}{\textbf{Normalized Effort Cost} $\downarrow$} & Trot & $1.5\!\pm\!0.2$ & $1.3\!\pm\!0.2$ & \textbf{1.0$\!\pm\!$0.2} \\
         & Flytrot & $1.9\!\pm\!0.3$ & $1.5\!\pm\!0.4$ & \textbf{1.0$\!\pm\!$0.2} \\
        \bottomrule
    \end{tabular}
\end{table}

We assessed each controller's performance using four metrics, logged in Table~\ref{tab:performance_comparison}. Success rate was determined based on three failure criteria: deviations in the base's height by more than 30\% from its desired value, orientation of the robot with respect to the ground normal dropping below 0.8 (indicating a significant tilt), and any infeasibility in solving the QP due to constraint violations. These thresholds were identified during hardware tuning as points where the robot physically fell. The slippage ratio was calculated as \((f_x^2 + f_y^2)/f_z^2\), where \(f_x\), \(f_y\), and \(f_z\) represent the ground reaction force components. This ratio represents the degree to which the friction cone was saturated, with higher values indicating worse performance. For the normalized costs, CCMPC was used as the baseline and normalized to 1. The other methods were compared relative to it in two categories: normalized tracking cost, where a higher value indicates poorer tracking of the desired CoM trajectory, and normalized effort cost, where a higher value indicates increased actuation effort. To ensure a fair comparison, the normalized cost and slippage ratio metrics were computed only for successful simulations. Failed simulations were excluded from these metrics to prevent skewed results due to failure conditions.

\begin{figure}[t]
     \centering
     
     % LMPC images (top row)
     \begin{subfigure}[t]{0.85\linewidth}
         \centering
         \includegraphics[width=0.2\textwidth]{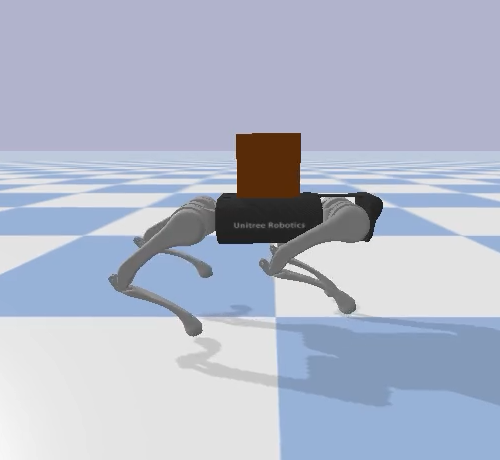}
         \includegraphics[width=0.2\textwidth]{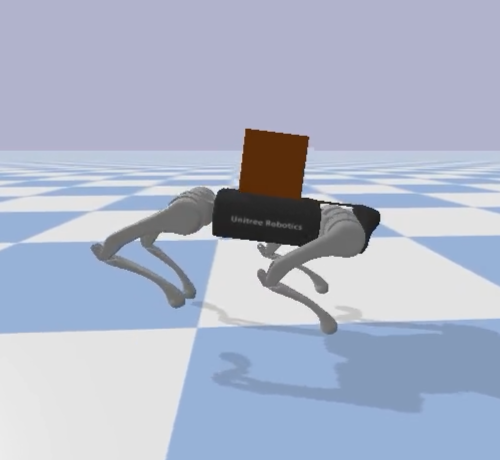}
         \includegraphics[width=0.2\textwidth]{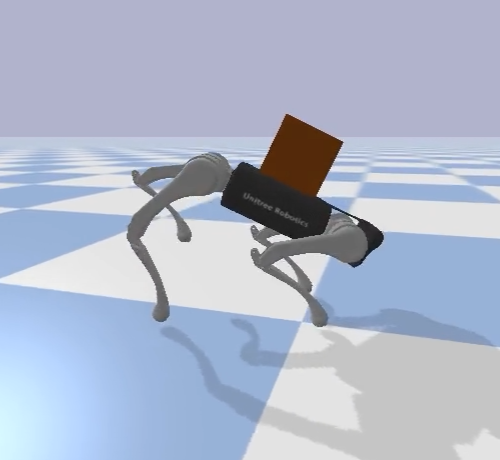}
         \includegraphics[width=0.2\textwidth]{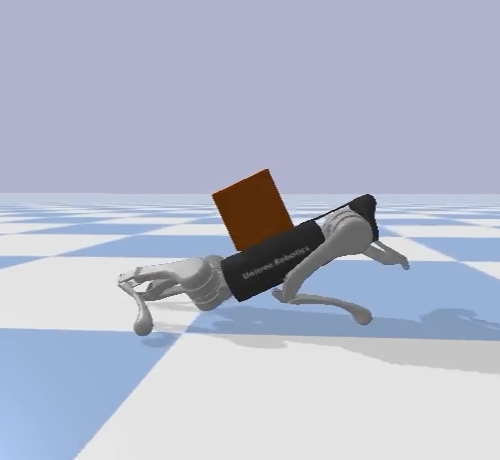}
         \label{fig:lmpc_tracking_sim}
        \includegraphics[width=\linewidth]{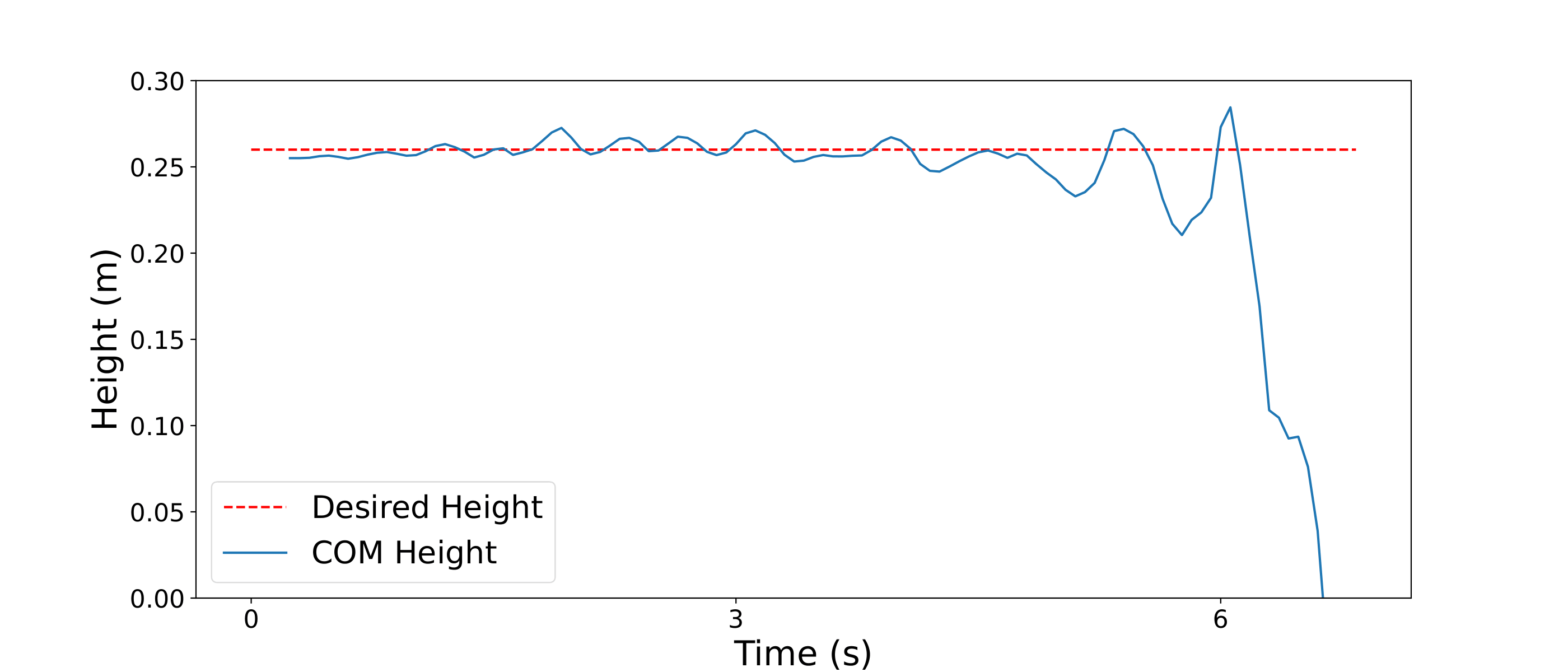}
         \caption{LMPC}
     \end{subfigure}
     % CCMPC images (second row)
     \begin{subfigure}[t]{0.85\linewidth}
         \centering
         \includegraphics[width=0.2\textwidth]{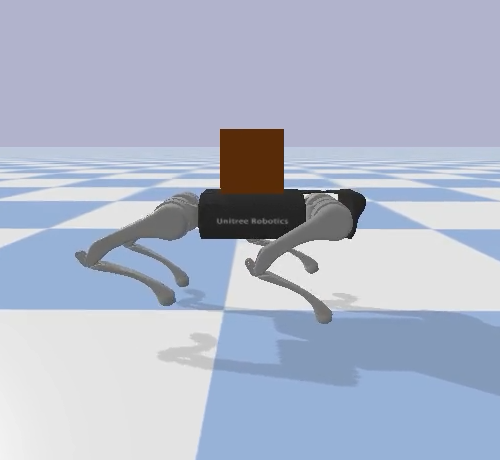}
         \includegraphics[width=0.2\textwidth]{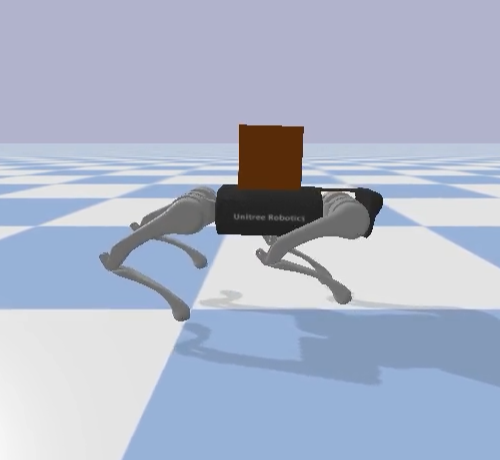}
         \includegraphics[width=0.2\textwidth]{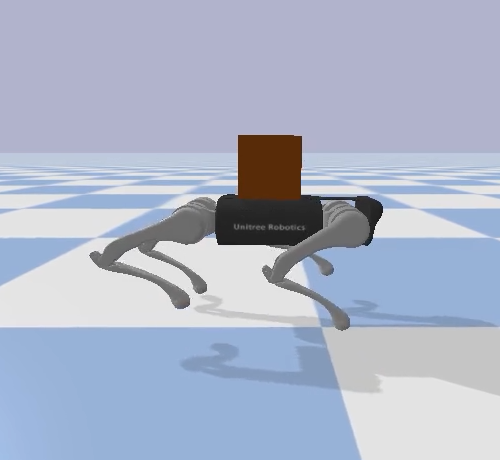}
         \includegraphics[width=0.2\textwidth]{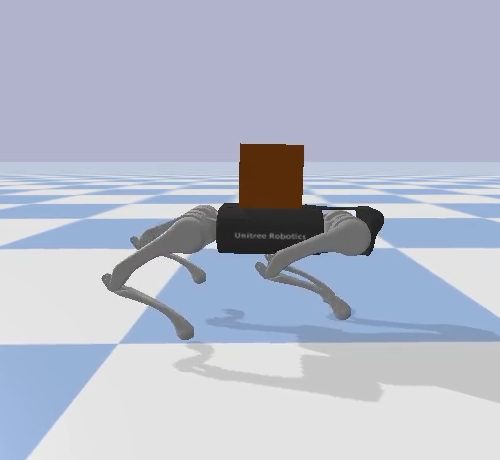}
         \label{fig:ccmpc_tracking_sim}
        \includegraphics[width=\linewidth]{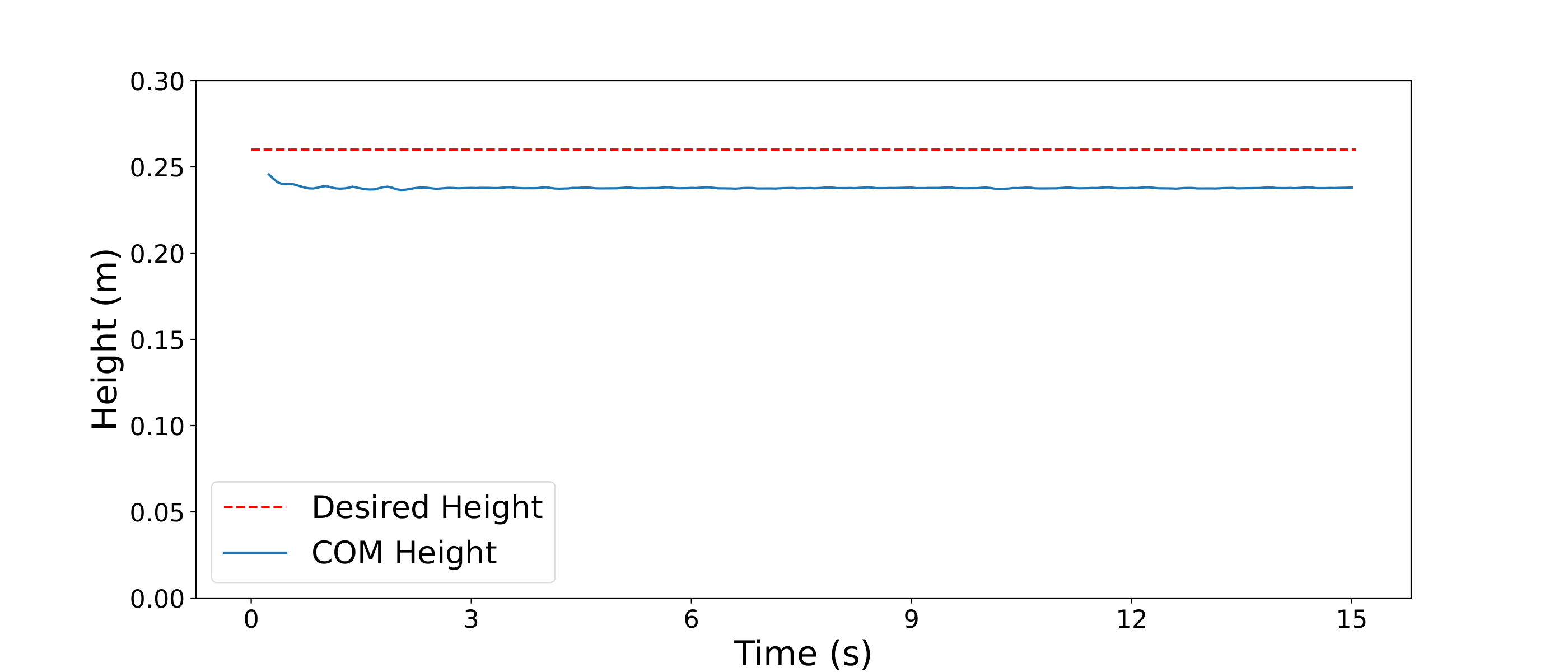}
         \caption{CCMPC}
     \end{subfigure}

     \caption{Simulated height tracking performance with an unmodeled 6 kg payload.}
     \label{fig:comparison_tracking_sim}
     \vspace{-3mm}
\end{figure}

% Fig 6

%%%%%%%%%%%%%%%%%%%%%%%%%%
\begin{figure*}
    \centering
    \begin{subfigure}[t]{0.18\textwidth}
        \centering
        \includegraphics[height=0.69\textwidth,width=\textwidth]
        {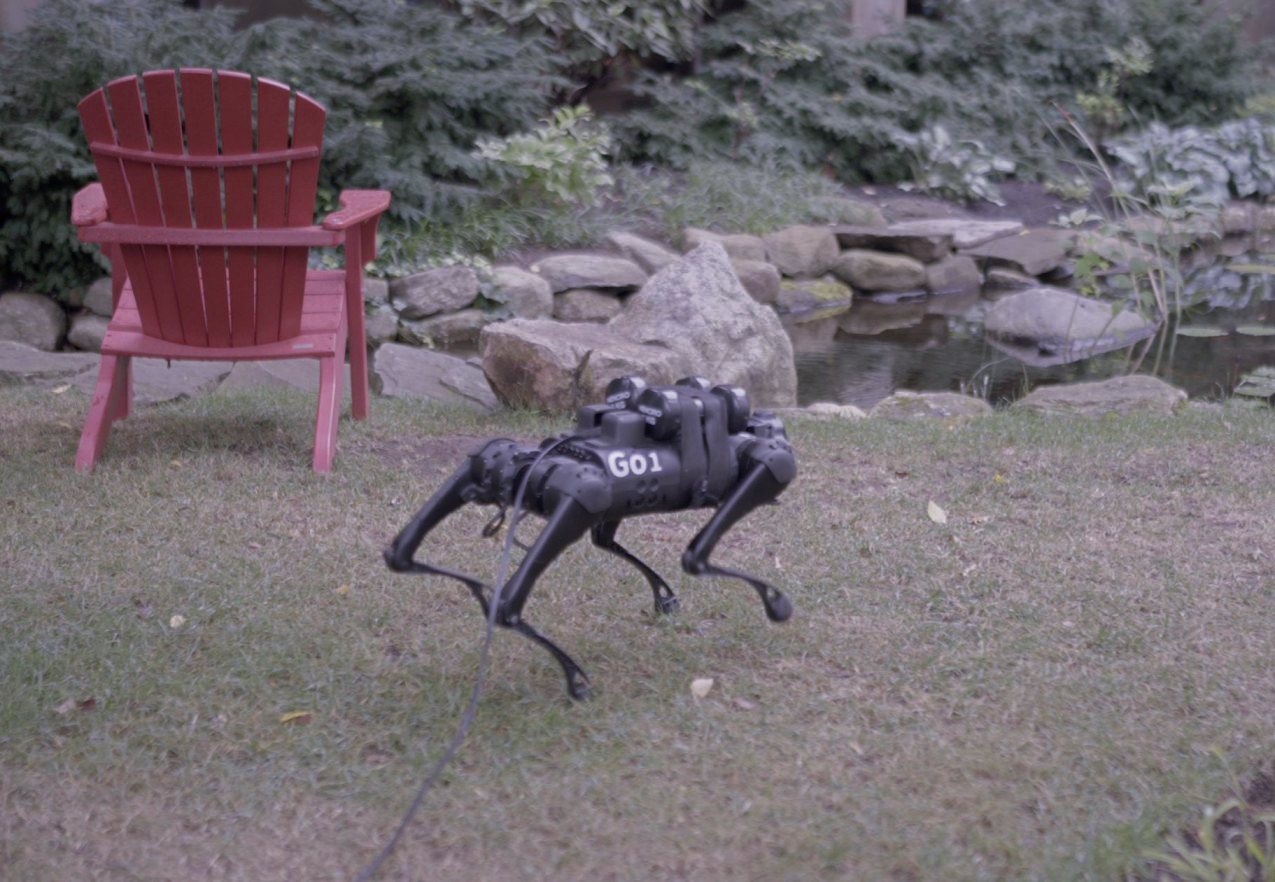}
        \caption{Grass}
    \end{subfigure}
    \begin{subfigure}[t]{0.18\textwidth}
        \centering
        \includegraphics[width=\textwidth]
        {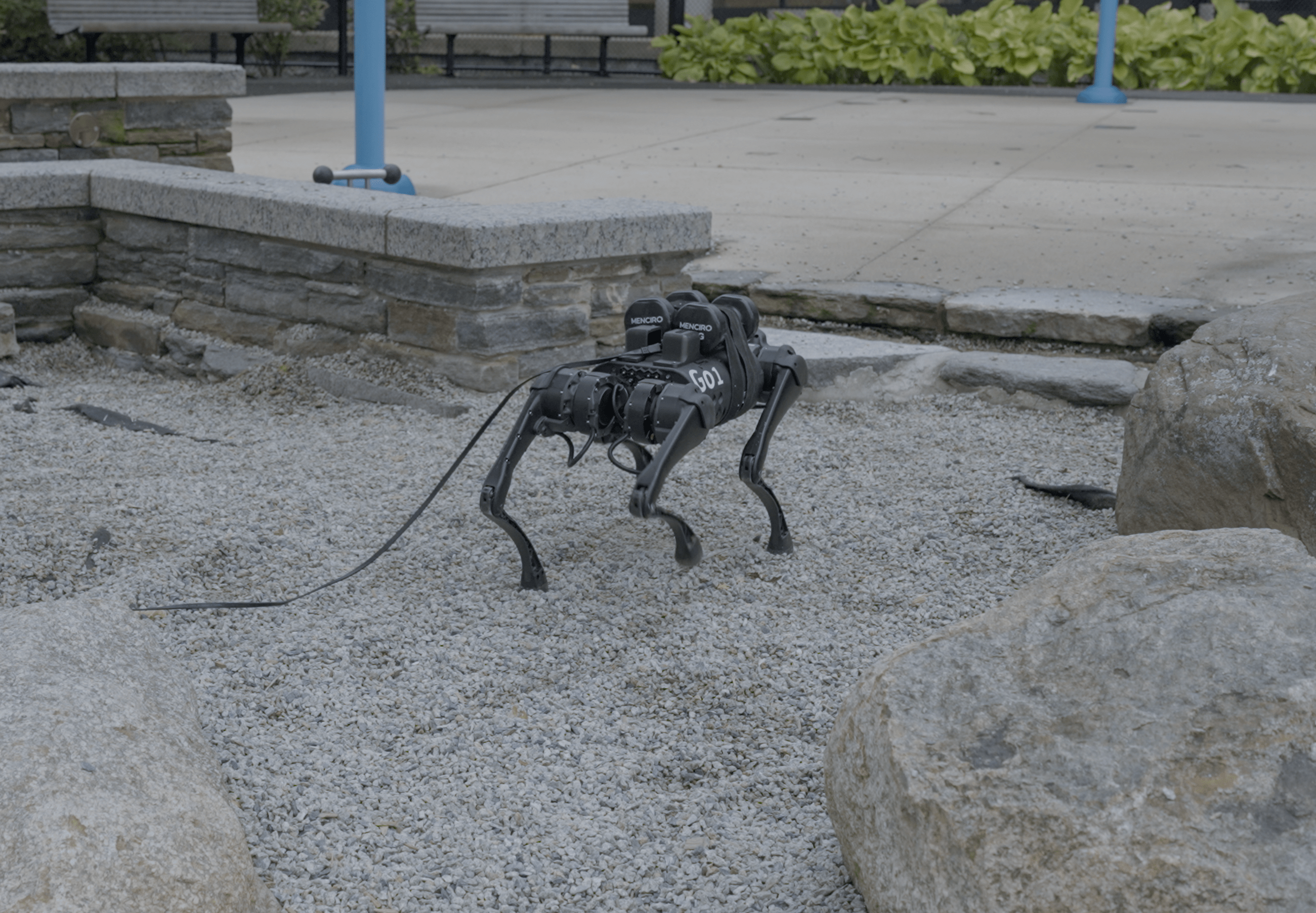}
        \caption{Gravel}
    \end{subfigure}
    \begin{subfigure}[t]{0.18\textwidth}
        \centering
        \includegraphics[width=\textwidth]
        {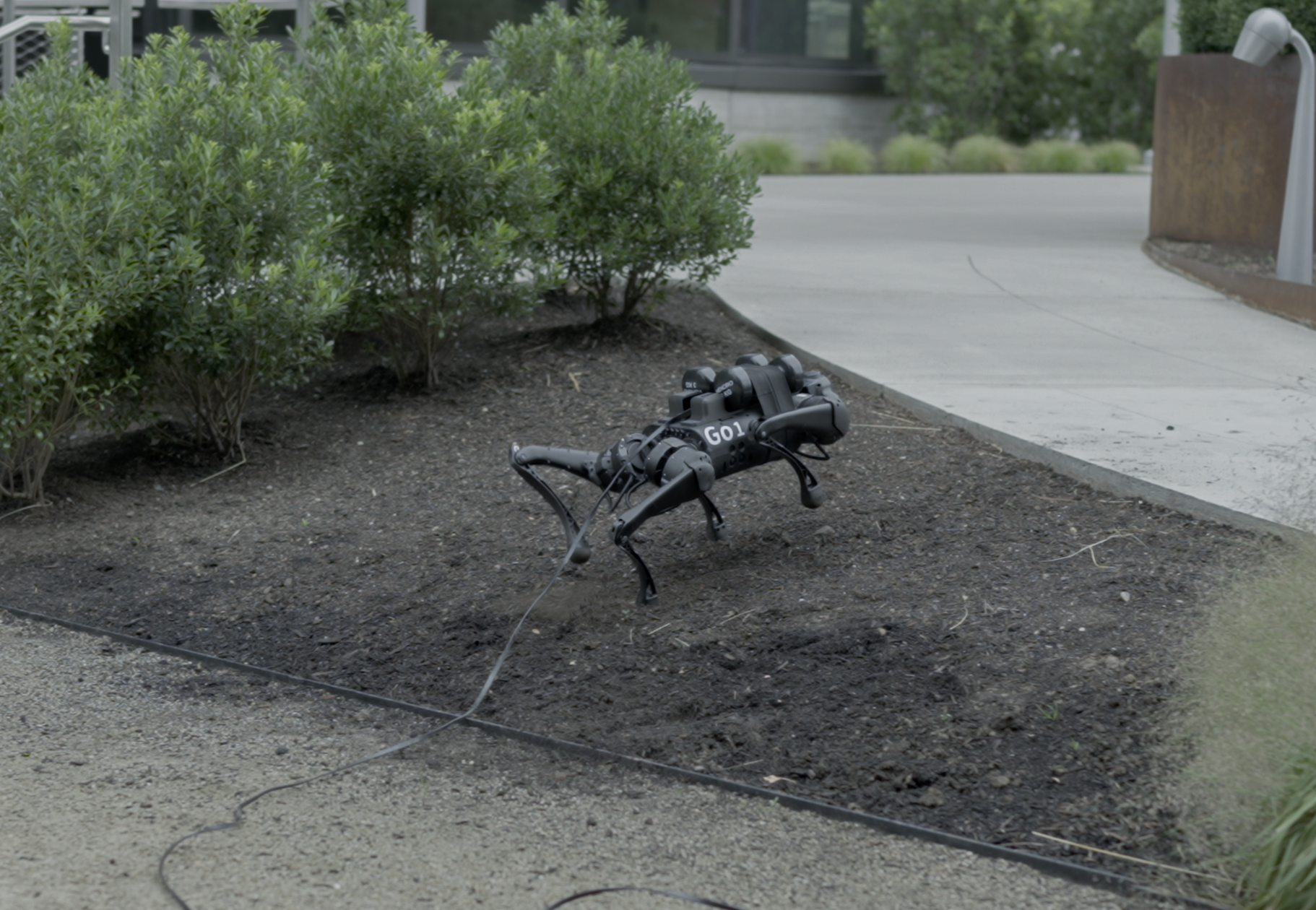}
        \caption{Muddy Slope}
    \end{subfigure}
    \begin{subfigure}[t]{0.18\textwidth}
        \centering
        \includegraphics[width=\textwidth]
        {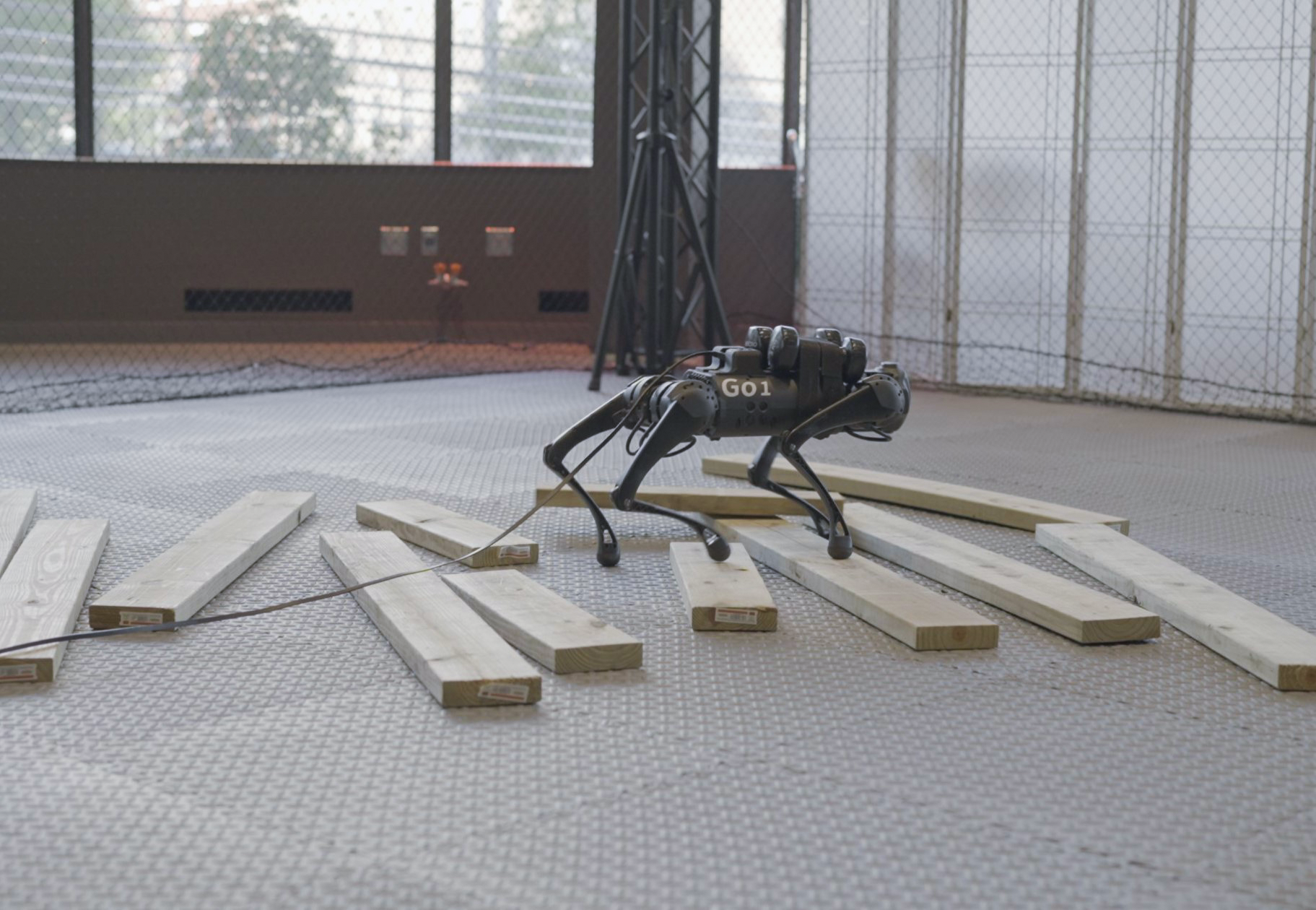}
        \caption{Planks}
    \end{subfigure}
    \begin{subfigure}[t]{0.18\textwidth}
        \centering
        \includegraphics[width=\textwidth]
        {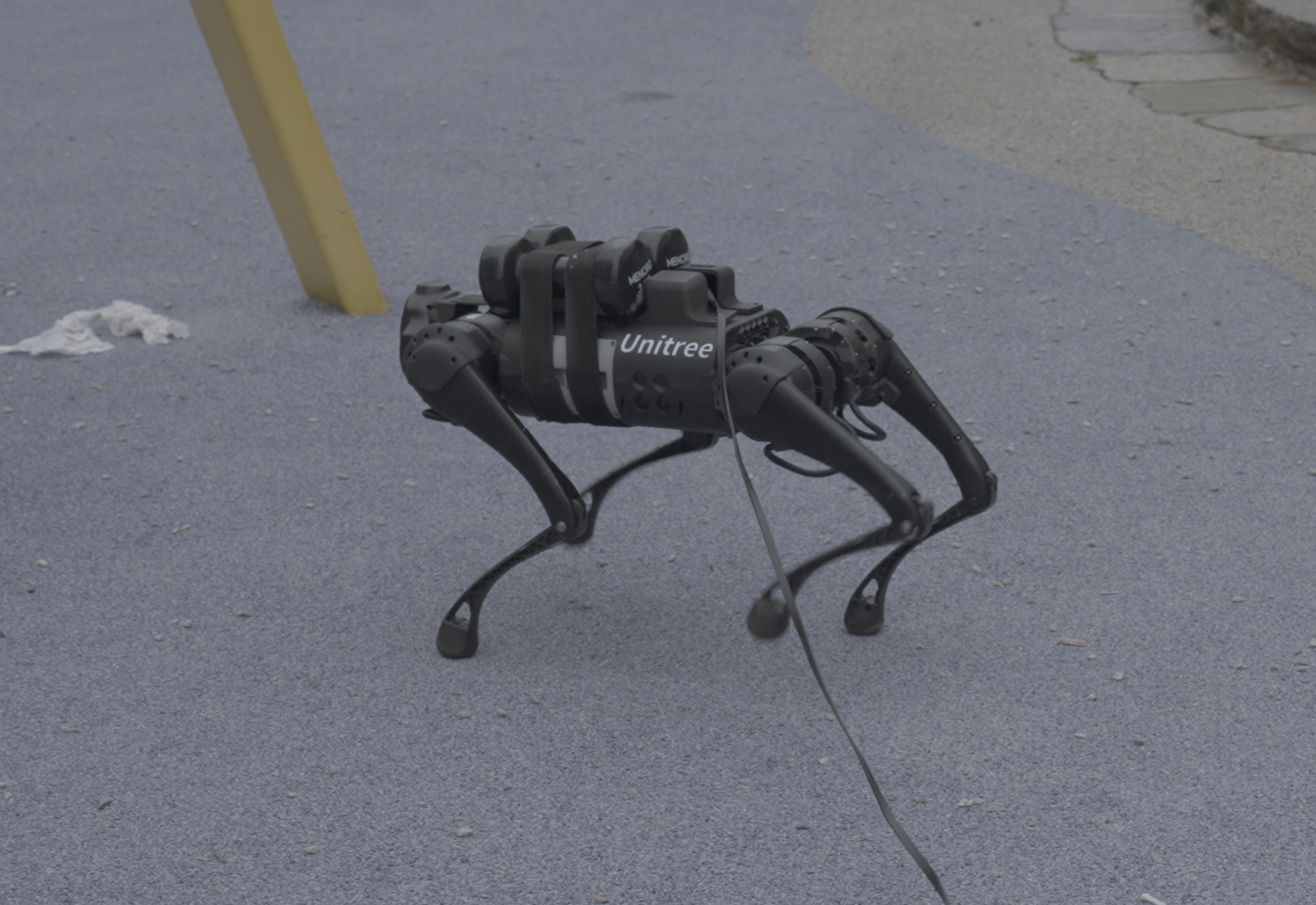}
        \caption{Bouncy Playground}
    \end{subfigure}
    \centering
    \begin{subfigure}[t]{0.18\textwidth}
        \centering
        \includegraphics[width=\textwidth, height=0.70\linewidth]
        {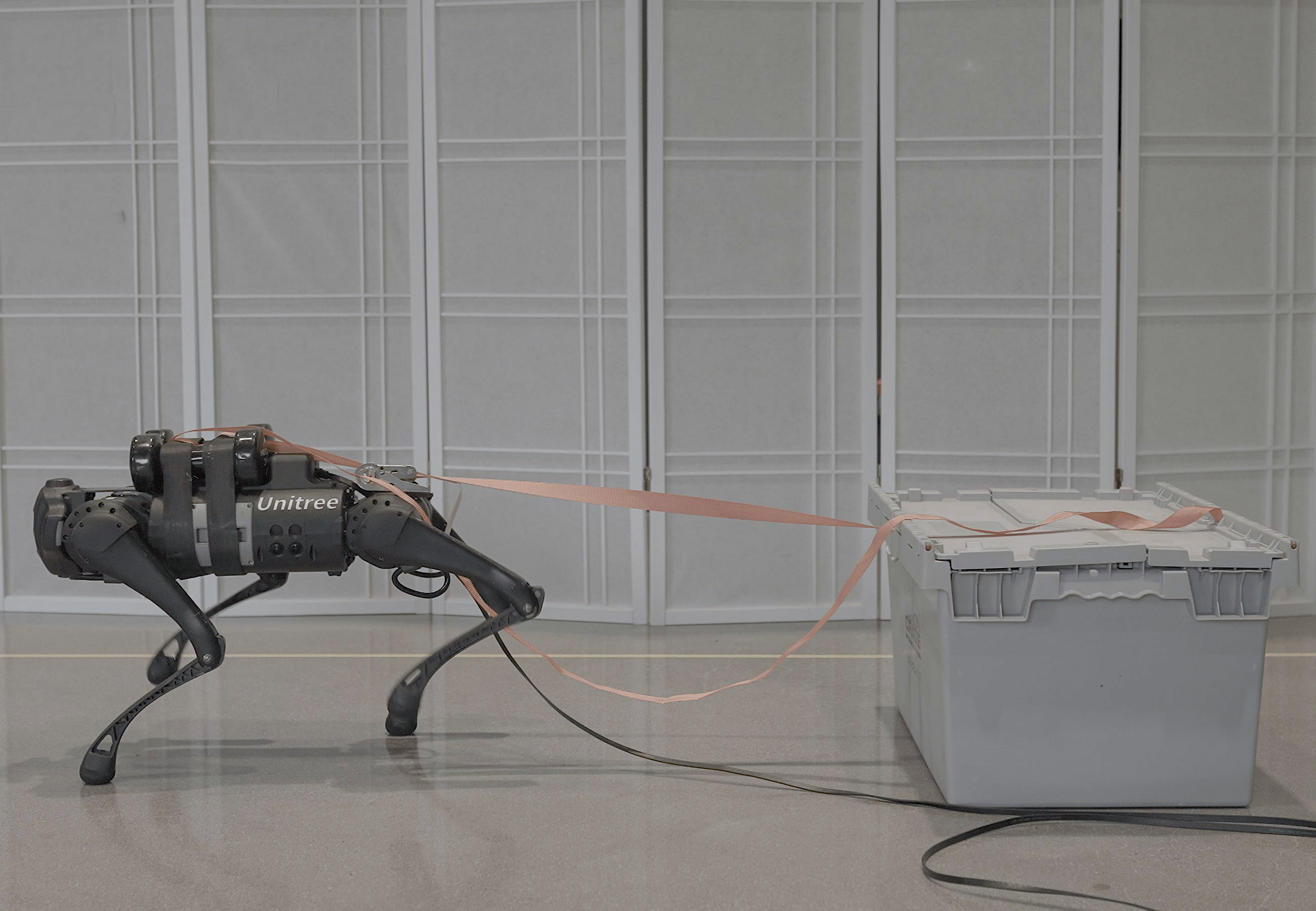}
        \caption{Pull Task}
    \end{subfigure}
    \begin{subfigure}[t]{0.18\textwidth}
        \centering
        \includegraphics[width=\textwidth, height=0.70\linewidth]
        {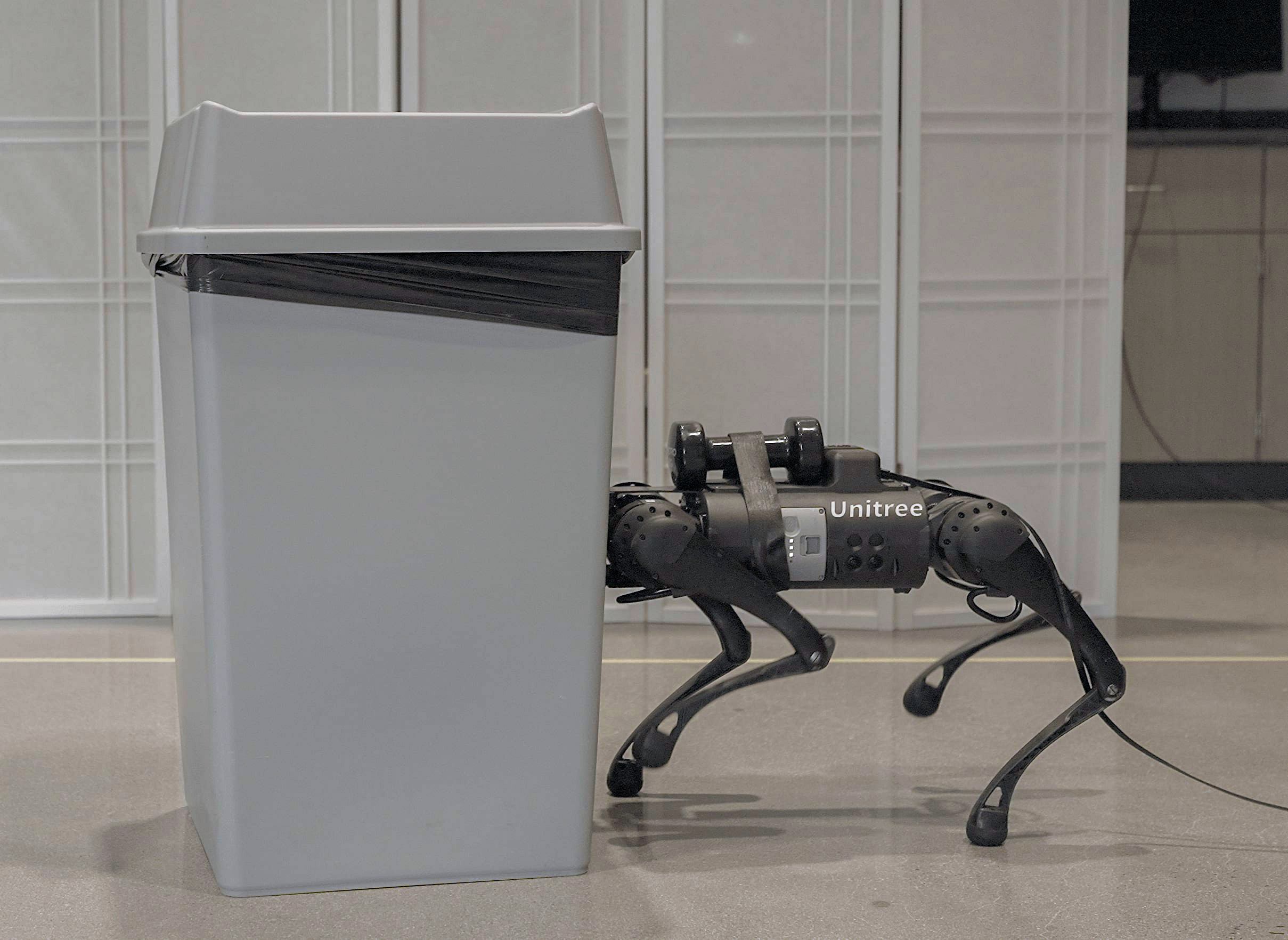}
        \caption{Push Task}
    \end{subfigure}
    \begin{subfigure}[t]{0.18\textwidth}
        \centering
        \includegraphics[width=\textwidth, height=0.70\linewidth]
        {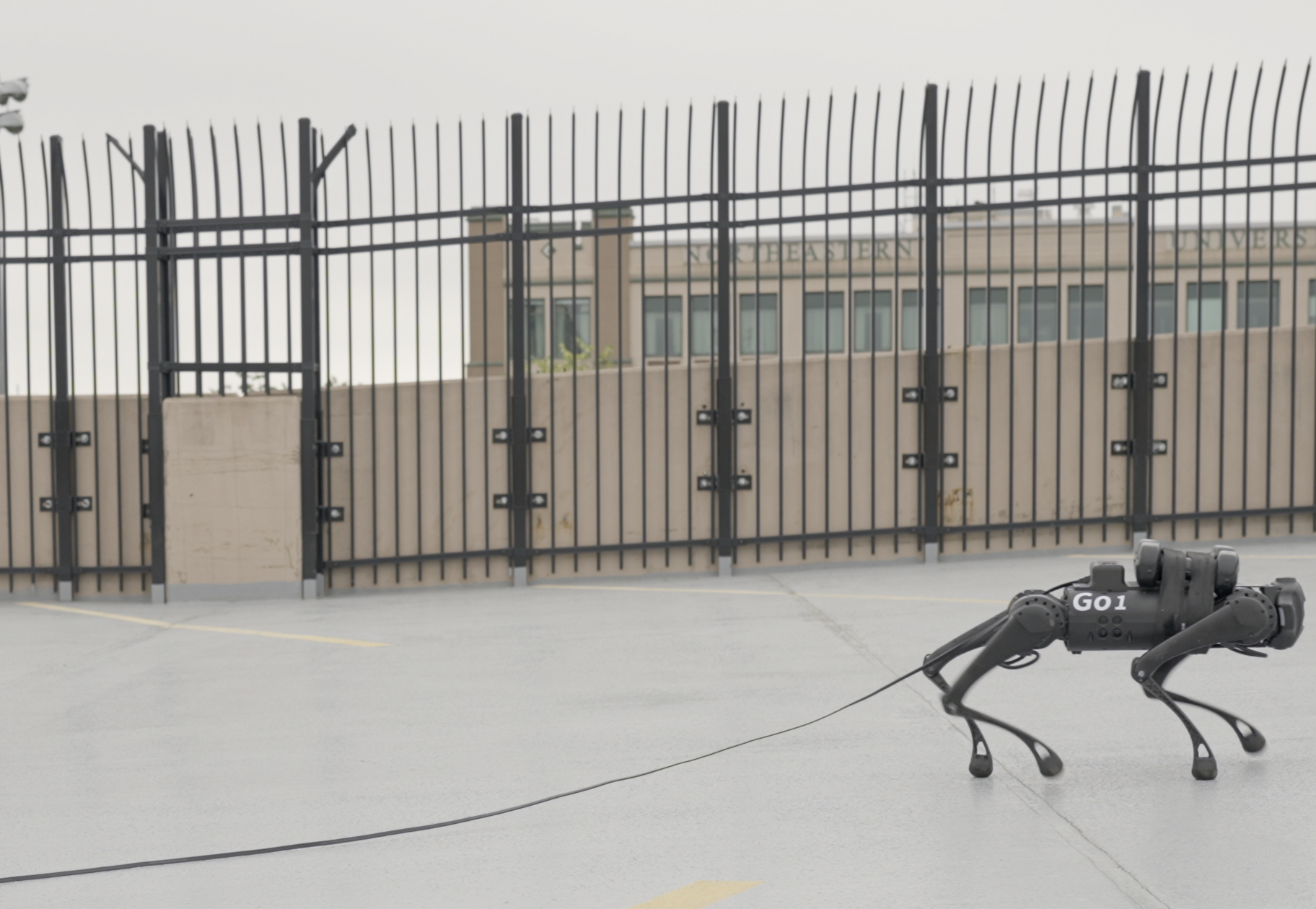}
        \caption{Concrete Slope}
    \end{subfigure}
    \begin{subfigure}[t]{0.18\textwidth}
        \centering
        \includegraphics[width=\textwidth, height=0.70\linewidth]
        {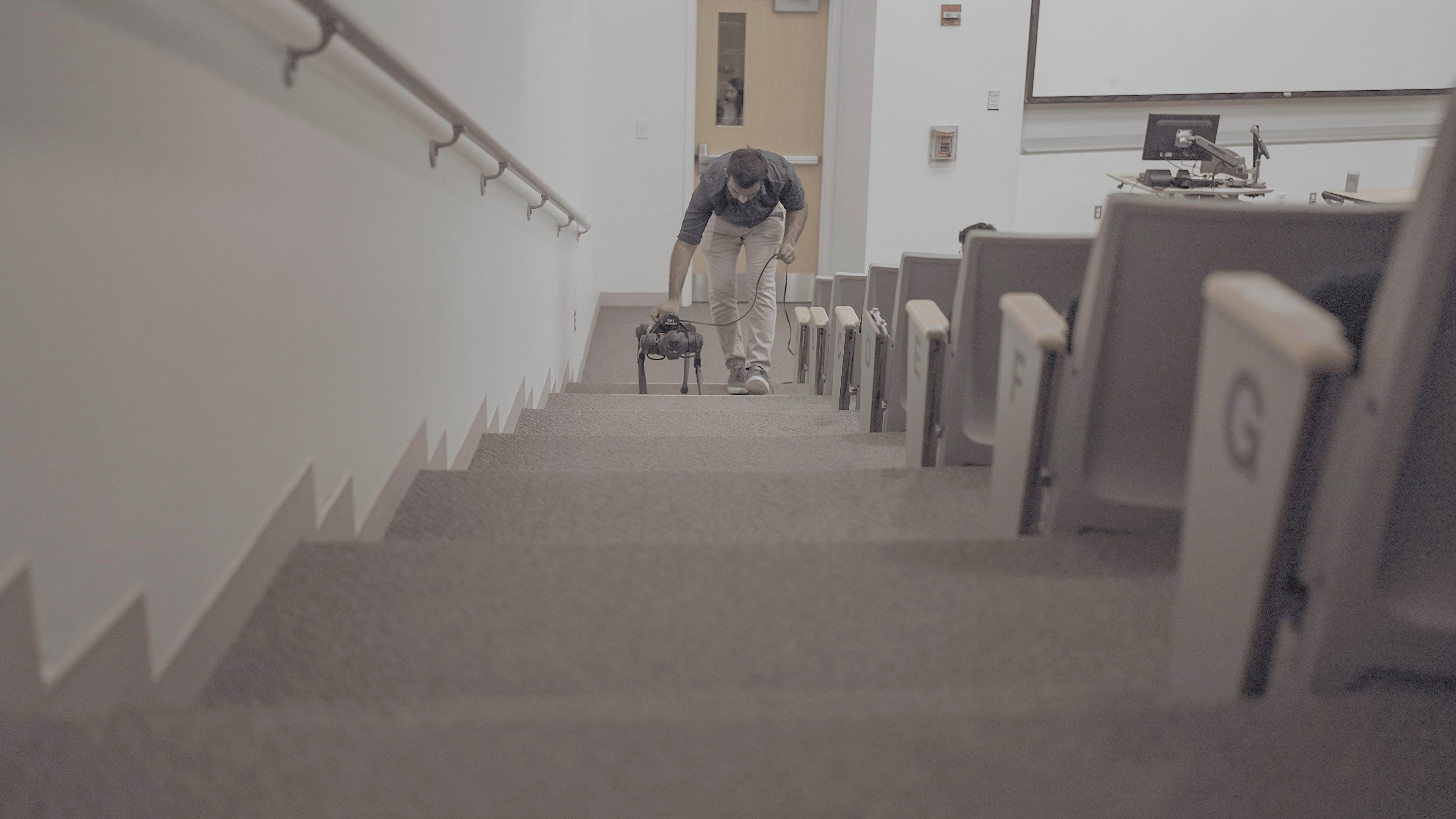}
        \caption{Blind Stair Climbing}
    \end{subfigure}
    \begin{subfigure}[t]{0.18\textwidth}
        \centering
        \includegraphics[width=\textwidth, height=0.70\linewidth]
        {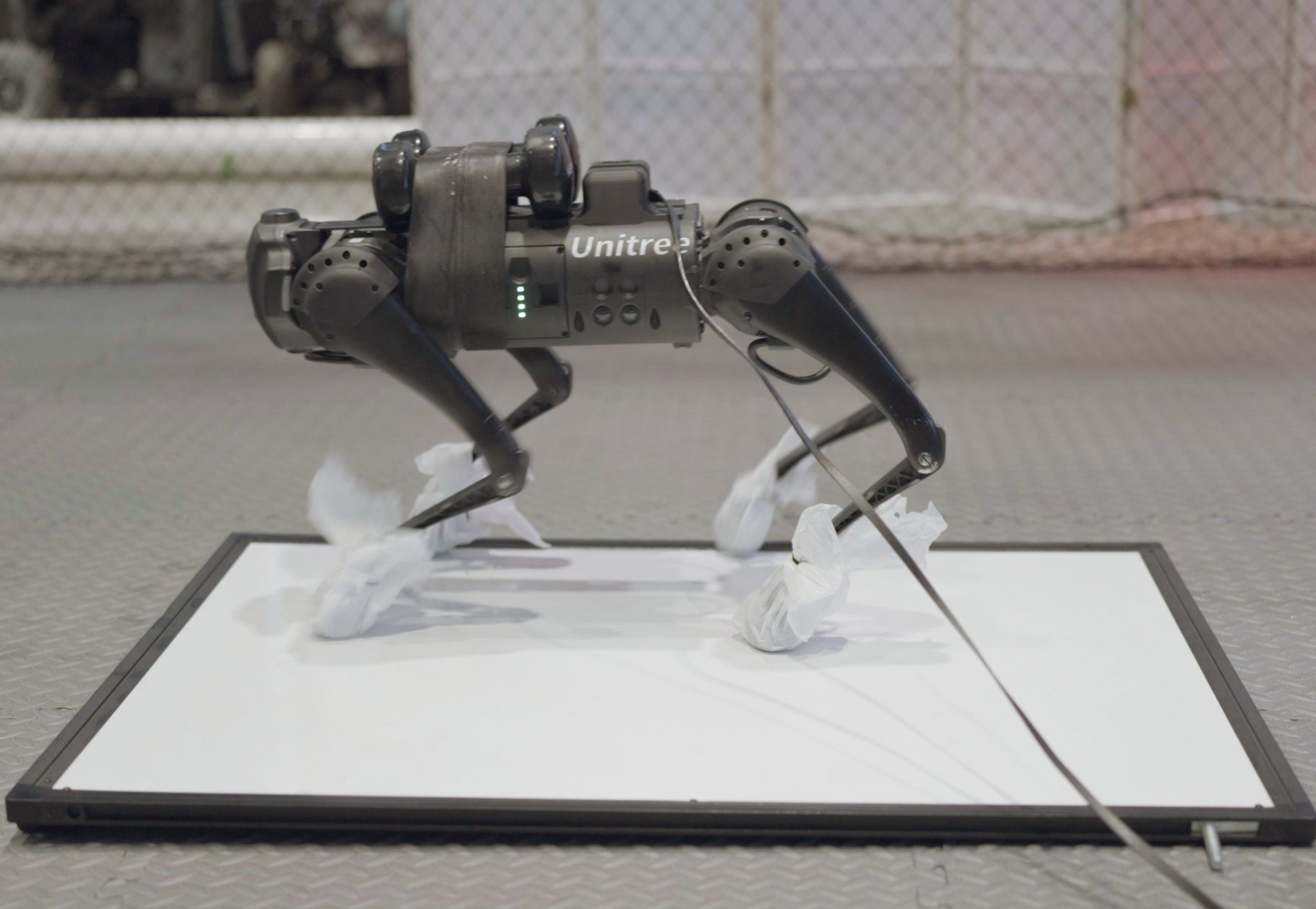}
        \caption{Slippery Whiteboard}
    \end{subfigure}
\caption{Hardware validation of CCMPC across different locomotion tasks and terrains. Video demonstrations can be found at \href{https://cc-mpc.github.io/}{https://cc-mpc.github.io/}.}
\label{fig:hardware_diversity}
\vspace{-1.1mm}
\end{figure*}
%%%%%%%%%%%%%%%%%%%%%%%%%%%

% FIGURE 7 [BEGIN]

\begin{figure}[t]
     \centering
     \begin{subfigure}[t]{0.84\linewidth}
         \centering
         \includegraphics[width=\linewidth]{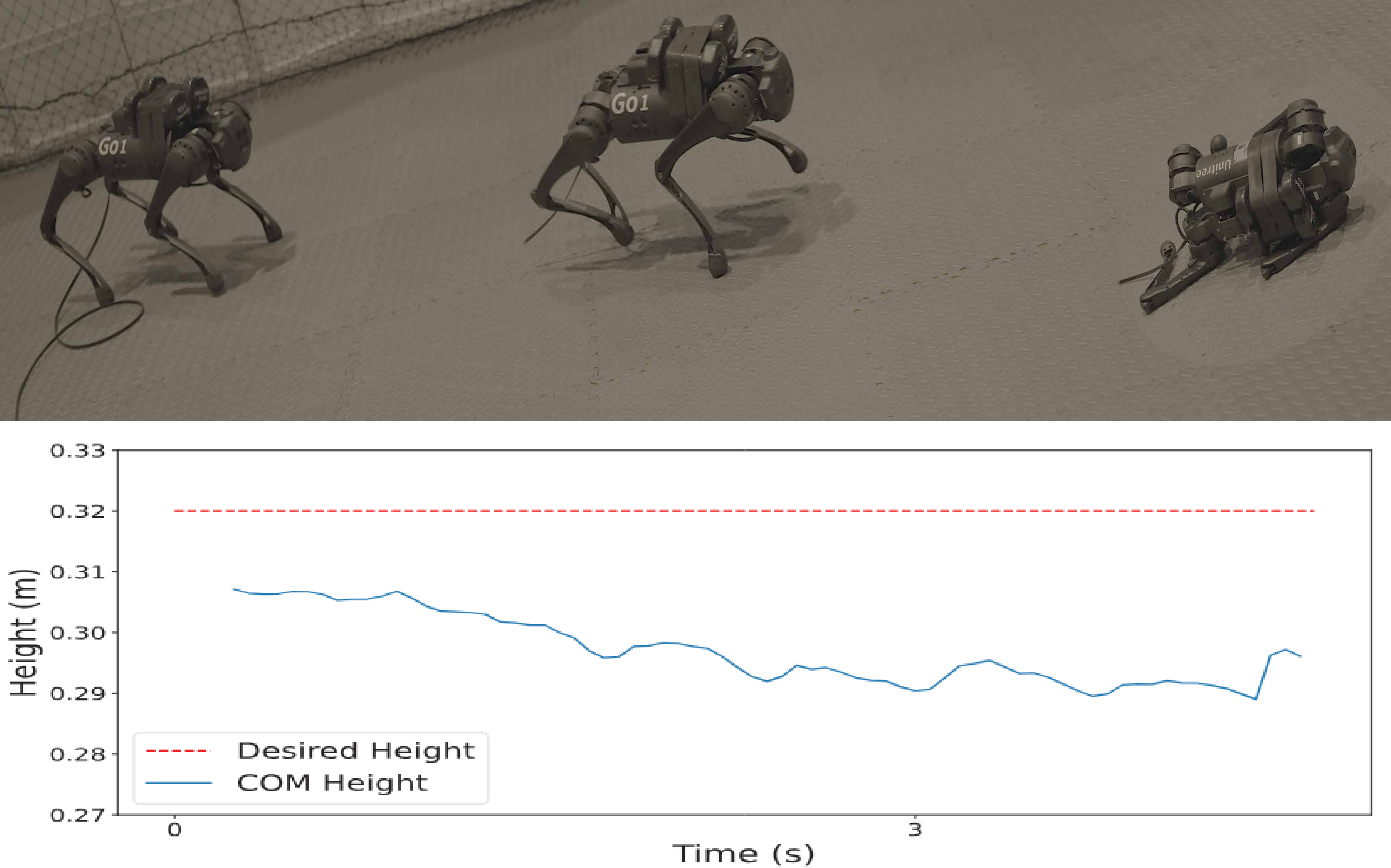}
         \caption{LMPC}
         \label{fig:lmpc_hardware_height_tracking}
     \end{subfigure}
     \begin{subfigure}{0.84\linewidth}
         \centering
         \includegraphics[width=\linewidth]{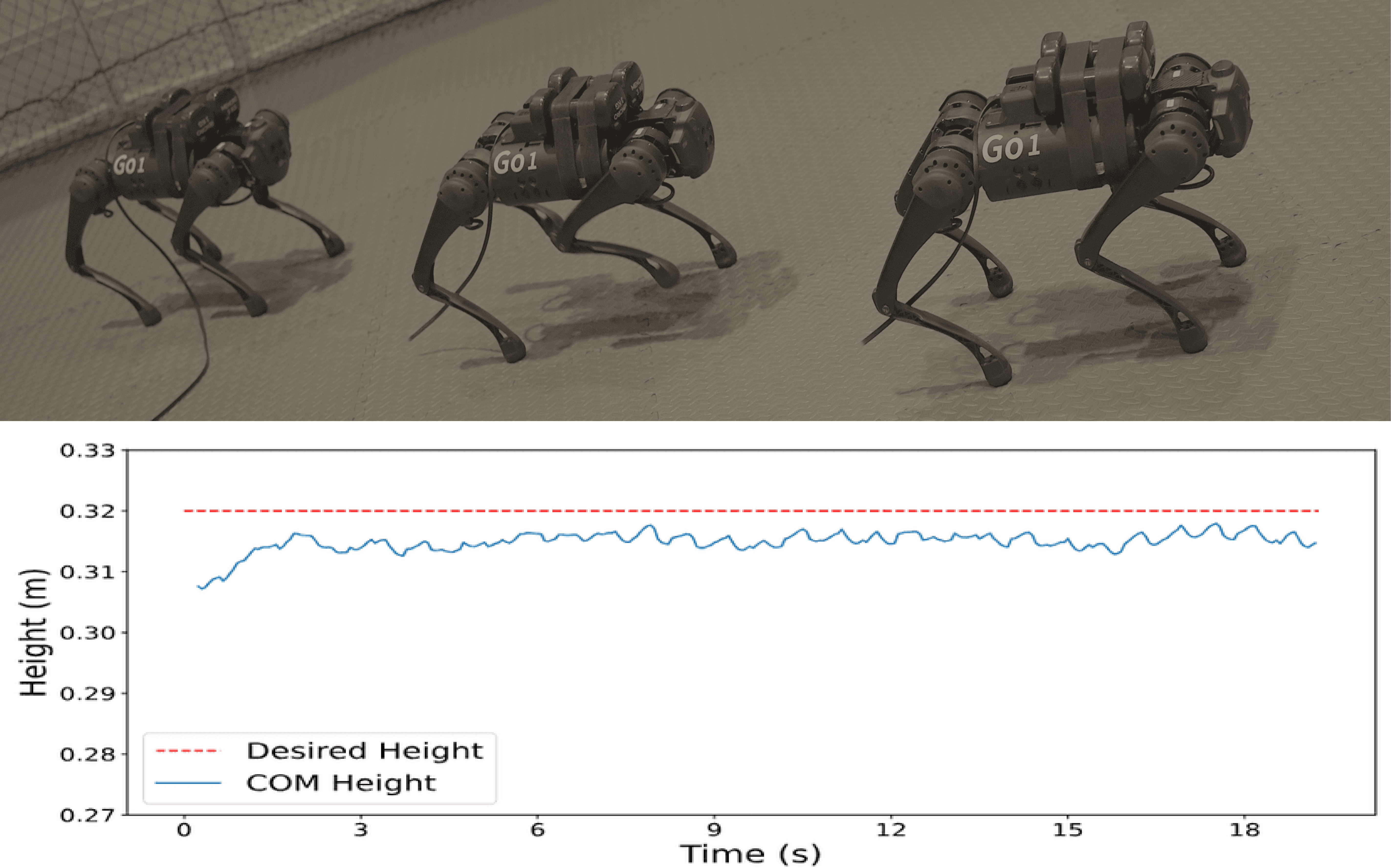}
         \caption{CCMPC}
         \label{fig:}
     \end{subfigure}
 \caption{Hardware height tracking performance with an unmodeled 6 kg payload.}
 \label{fig:ccmpc_hardware_height_tracking}
 \vspace{-2.0mm}
\end{figure}

% FIGURE 4 [END]

Our quantitative results show that LMPC had the lowest success rate, highest slippage ratio, and highest normalized cost. HMPC showed improvements over LMPC due to hand-tuned constraint tightening. However, as HMPC maintains a constant level of constraint tightening throughout the MPC horizon, it was less effective in adapting to disturbances, like unmodeled payloads and uneven planks. CCMPC obtained the best performance across both gaits, dynamically adjusting constraints in real-time to balance safety and tracking, thereby minimizing slippage and maintaining stability~\cite{related_works_gazar_wholebody}.

Fig.~\ref{fig:comparison_tracking_sim} illustrates the height tracking performance, where unlike CCMPC, LMPC struggles to maintain stability and results in a fall. The slight offset in CoM height under CCMPC arises from prioritizing safety constraints, such as preventing foot slippage, allowing minor deviations in height to ensure robust performance across various conditions.

Fig.~\ref{fig:sim_hard} demonstrates simulations on various challenging terrains, where the robot traversed random elevation wavefield, randomly placed planks, and performed blind stair climbing. High-speed tests at 1.75 m/s were also conducted using the flytrot gait. In all these scenarios, the robot performed blind locomotion while carrying a fixed 6 kg payload. Both LMPC and HMPC struggled to maintain stability, while CCMPC successfully maintained stability and tracking, effectively handling unmodeled terrain and payload uncertainties.

\subsection{Hardware Validation}

To validate our approach in real-world conditions, we conducted hardware experiments using the Unitree Go1 robot, replicating key scenarios from the simulations. The first experiment, illustrated in Fig.~\ref{fig:visual_only_hardware_comparison}, involved loading the robot with a 6 kg payload and commanding it to walk over randomly placed wooden planks at a speed of 0.5 m/s. This experiment was conducted under blind locomotion conditions, where the controller was unaware of the planks or additional mass. While LMPC failed to navigate this terrain, CCMPC successfully guided the robot across the planks. 

We explored the versatility of our control algorithm through various challenging terrain experiments, as shown in Fig.~\ref{fig:hardware_diversity}. The robot, loaded with additional unknown mass, successfully climbed stairs and navigated muddy slopes, grass, and gravel. The robot also performed tasks on objects, such as pushing and pulling an unknown 5 kg payload. In another experiment, the robot walked over a whiteboard coated with cooking oil, significantly reducing the friction between the robot's feet and the surface. For this test, we reduced the robot's speed to 0.1 m/s and adjusted the coefficient of friction in the MPC from 0.4 to 0.2 to account for the slippery surface. Despite these adverse conditions, the robot successfully navigated all these surfaces, showcasing its ability to handle diverse ground textures and maintain stability on slippery and uneven surfaces.

To demonstrate the repeatability of our control strategy, we conducted a series of progressive load tests, as shown in Fig.~\ref{fig:ccmpc_hardware_height_tracking}. In these experiments, the robot was commanded to move forward at a speed of 0.25 m/s while carrying increasing payloads: 1.3 kg, 2.2 kg, 3.0 kg, 4.3 kg, 6.0 kg, and 7.3 kg. The robot traversed flat ground during all tests. The results showed that LMPC maintained stability up to 4.3 kg but failed to navigate with a 6.0 kg payload. In contrast, CCMPC successfully handled both 6.0 kg and 7.3 kg payloads, even exceeding the manufacturer's recommended maximum payload of 5.0 kg. More hardware demonstrations can be found in the supplementary video, which provides a comprehensive overview of the robot's performance across these varied conditions. %These experiments highlight the effectiveness of our CCMPC algorithm in handling a wide range of scenarios and maintaining stable locomotion across various conditions.

\section{Conclusion}
In this paper, we presented a Chance-Constrained Model Predictive Control (CCMPC) framework to generate optimal ground reaction forces for quadrupedal locomotion. CCMPC generates a unified control policy to stabilize quadrupedal locomotion across a wide range of unmodeled payloads and varying terrain conditions in real-time with minimal parameter tuning. The effectiveness of our approach was validated through extensive simulation and hardware experiments. 

However, some limitations were observed, particularly with heavier payloads. Tightening control constraints in anticipation of added mass can lead to QP infeasibility. To address this, future work will explore machine learning methods~\cite{icaa,icra} to actively learn payload dynamics. Additionally, we aim to extend this approach to loco-manipulation tasks in bipedal robots handling unknown payloads~\cite{locomanipulation}.

\section{Acknowledgments}
This work is partially supported by the Impact Engines program at Northeastern University.

\bibliographystyle{IEEEtran}
\bibliography{IEEEabrv,references}

\end{document}